\documentclass[10pt,twocolumn,letterpaper]{article}

\usepackage{cvpr}              

\usepackage{multirow}

\usepackage{fancyvrb,newverbs}

\definecolor{cverbbg}{gray}{0.93}

\newenvironment{lcverbatim}
 {\SaveVerbatim{cverb}}
 {\endSaveVerbatim
  \flushleft\fboxrule=0pt\fboxsep=.5em
  \colorbox{cverbbg}{%
    \makebox[\dimexpr\linewidth-2\fboxsep][l]{\BUseVerbatim{cverb}}%
  }
  \endflushleft
}

\newverbcommand{\cverb}
  {\setbox\verbbox\hbox\bgroup}
  {\egroup\colorbox{cverbbg}{\box\verbbox}}

\definecolor{cvprblue}{rgb}{0.21,0.49,0.74}
\usepackage[pagebackref,breaklinks,colorlinks,allcolors=cvprblue]{hyperref}

\def\paperID{10288} 
\def\confName{CVPR}
\def\confYear{2025}

\title{PQPP: A Joint Benchmark for Text-to-Image\\ Prompt and Query Performance Prediction}

\author{Eduard Poesina$^{1,\diamond}$, {Adriana Valentina} Costache$^{1,\diamond}$, Adrian-Gabriel Chifu$^2$,\\
Josiane Mothe$^3$, {Radu Tudor} Ionescu$^{1,}$\thanks{Corresp. author: {\tt\scriptsize raducu.ionescu@gmail.com}. ~$^\diamond$Equal contrib.}\\
$^1$University of Bucharest, $^2$Aix-Marseille Universit\'{e}, $^3$Universit\'{e} Toulouse Jean-Jaur\`{e}s
}

\begin{document}
\maketitle

\begin{abstract}
Text-to-image generation has recently emerged as a viable alternative to text-to-image retrieval, driven by the visually impressive results of generative diffusion models. Although query performance prediction is an active research topic in information retrieval, to the best of our knowledge, there is no prior study that analyzes the difficulty of queries (referred to as \emph{prompts}) in text-to-image generation, based on human judgments. To this end, we introduce the first dataset of prompts which are manually annotated in terms of image generation performance. Additionally, we extend these evaluations to text-to-image retrieval by collecting manual annotations that represent retrieval performance. We thus establish the first joint benchmark for prompt and query performance prediction (PQPP) across both tasks, comprising over 10K queries. Our benchmark enables (i) the comparative assessment of prompt/query difficulty in both image generation and image retrieval, and (ii) the evaluation of prompt/query performance predictors addressing both generation and retrieval. We evaluate several pre- and post-generation/retrieval performance predictors, thus providing competitive baselines for future research. Our benchmark and code are publicly available at \url{https://github.com/Eduard6421/PQPP}.
\vspace{-0.2cm}
\end{abstract}

\vspace{-0.2cm}
\section{Introduction}
\vspace{-0.1cm}

\begin{figure*}[t]
  \centering
  \includegraphics[width=0.855\linewidth]{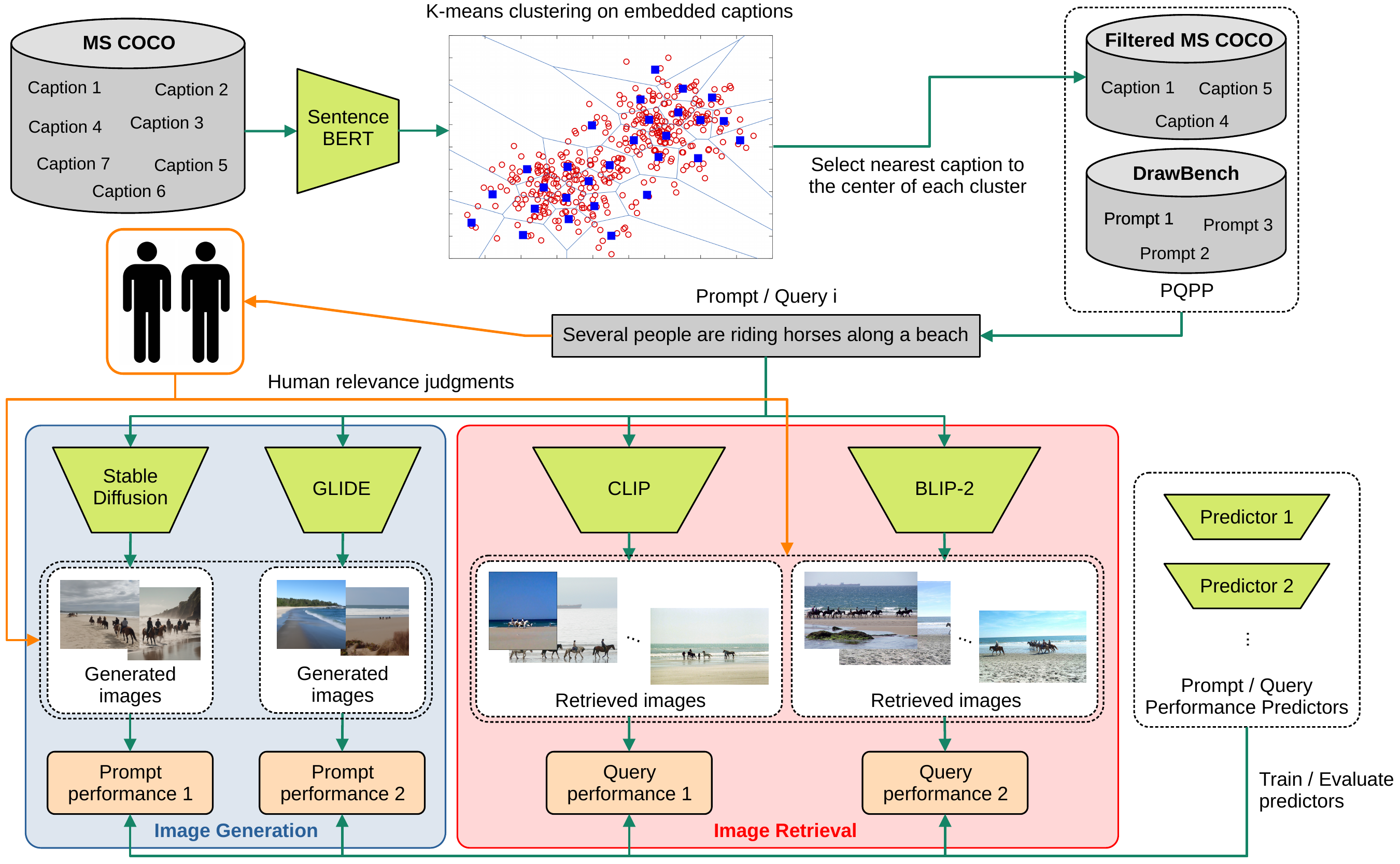}
  \vspace{-0.25cm}
  \caption{We select a set of 10K captions from MS COCO \citep{Lin-ECCV-2014} using k-means clustering, which are further merged with prompts from DrawBench \cite{Saharia-NeurIPS-2022}. Next, we collect human relevance judgments in two scenarios: image generation and image retrieval. For each prompt/query, we generate images with two diffusion models (Stable Diffusion XL~\citep{Podell-ICLR-2024} and GLIDE~\citep{Nichol-ICML-2021}) and retrieve images from MS COCO with two vision-language models (CLIP~\citep{Radford-ICML-2021} and BLIP-2~\citep{Li-ICML-2023}). Based on the collected relevance judgments, we score each prompt/query in terms of generation and retrieval performance, respectively. Finally, we train and evaluate multiple prompt/query performance predictors on the proposed benchmark. Best viewed in color.}
  \label{fig_pipeline}
  \vspace{-0.3cm}
\end{figure*}

In recent years, more and more people embraced the use of large language models (LLMs) instead of traditional search engines \citep{Ai-AIO-2023,Caramancion-ArXiv-2024,zhu2024large}. The advent of generative diffusion models \citep{Croitoru-TPAMI-2023,Dhariwal-NeurIPS-2021,Nichol-ICML-2021,Rombach-CVPR-2022} capable of generating high-quality and realistic images triggered a similar trend in text-to-image retrieval. This paradigm shift in information retrieval (IR) calls for an extensive exploration of research topics related to generative models. 
One is query performance prediction (QPP), which refers to the process of estimating the performance level of an IR system for a given query~\citep{carmel2010estimating}, this being an active research topic in IR~\citep{datta2022deep,dejean2020forward,faggioli2023query,meng2023query,Poesina-SIGIR-2023}. In the context of generative AI for text-to-image generation, if a prompt is predicted to be difficult, the system could initiate a conversation to refine the prompt in order to overcome the difficulty and improve the final output. Moreover, the system could indicate to the user its inability to provide a satisfactory image, or it could give positive feedback to the user when the prompt is predicted as easy. In general, when a prompt/query is predicted as difficult for a generation/retrieval system, additional processes can be activated, such as the automatic reformulation, the allocation of extra resources, or the addition of pre- or post-processing. We discuss further applications and concrete use cases of prompt performance prediction in Appendix \ref{sec_usefulness}.

With the growing popularity of generative models and retrieval augmented generation, understanding prompt effectiveness in both generation and retrieval becomes increasingly essential. While text-to-image generation and retrieval share common elements, they differ fundamentally in terms of task requirements and success criteria. In text-to-image generation, prompts must guide models to produce images that not only capture the specified elements, but also meet aesthetic and contextual expectations. Conversely, in retrieval, the query role is to retrieve pre-existing images that match the query as closely as possible within a given dataset, where success depends on how well the query aligns with the available data. Studying both tasks together is critical. Examining prompt/query difficulty across generation and retrieval allows us to determine whether the factors that make a prompt challenging in generation (\eg~complexity, specificity, ambiguity) are the same as those in retrieval. 
To the best of our knowledge, there is no prior study that jointly analyzes the difficulty of prompts/queries in text-to-image generation and retrieval using human relevance judgments. In this study, we introduce the first dataset of prompts/queries which are manually annotated in terms of both image generation performance and image retrieval performance. Additionally, we provide baseline results with various pre- and post-performance predictors, setting a standard for future multimodal QPP research.

Our new Prompt and Query Performance Prediction (PQPP) benchmark comprises 10,200 text samples that serve as prompts for text-to-image generation and queries from text-to-image retrieval. To obtain representative prompts/queries for both generation and retrieval, we combine the 200 prompts from DrawBench \cite{Saharia-NeurIPS-2022} with 10,000 diverse captions from the Microsoft Common Objects in Context (MS COCO) dataset \citep{Lin-ECCV-2014}. While DrawBench is a recently proposed set of prompts that aims to test the capability of generative models in diverse scenarios, MS COCO is a well-established data collection that allows cross-task studies and is cost-effective.  MS COCO provides high-quality image annotations, as well as a wide range of everyday scenes and objects, being widely used across several computer vision tasks, \eg~object detection, object segmentation, and image captioning. Notably, with the advent of text-to-image generation, MS COCO has also been employed in many works for the text-to-image generation task \cite{Bao-CVPR-2023,Kang-CVPR-2023,Saharia-NeurIPS-2022,Xue-NeurIPS-2024}. 
To select a representative and diverse set of prompts from over 590K captions in MS COCO, we employ a k-means clustering algorithm on the embedding space of a sentence BERT model \citep{Thakur-NAACL-2020}, where $k\!=\!10,\!000$. 

To analyze prompt performance in both generation and retrieval, we examine the same set of 10,200 prompts across two generative models and two retrieval models (see Figure~\ref{fig_pipeline}).
For the generation task, we use SDXL \citep{Podell-ICLR-2024} and GLIDE \citep{Nichol-ICML-2021}, each generating two images per prompt. Both models are state-of-the-art diffusion models, yet they rely on different designs, which  leads to variability in terms of image quality and relevance.
Next, we collect over 247K relevance judgments from 147 human annotators for 40,800 images generated by the chosen diffusion models. 
For the retrieval task, we employ two state-of-the-art vision-language models, CLIP \citep{Radford-ICML-2021} and BLIP-2 \citep{Li-ICML-2023}, to retrieve images for the 10,200 queries. These models use different architectures and training data, impacting their retrieval performance for the same query. We start from preliminary relevance judgments determined via the pre-trained sentence BERT \citep{Thakur-NAACL-2020} applied on captions. If a caption from MS COCO is similar to a query in the BERT embedding space, the image corresponding to the respective caption is added to the set of images that are potentially relevant to the respective query. Next, we collect 1.39M ground-truth relevance judgments from 91 human evaluators to annotate the 10,200 queries. In summary, we collect over 1.6M annotations to estimate the difficulty score of each query included in PQPP, from both generation and retrieval perspectives.

We conduct preliminary experiments to compare prompt/query difficulty in image generation vs.~image retrieval. Our findings show that there is a very low correlation between the two tasks (see Section \ref{Sub:GvsR}), which justifies the need to study the novel task of prompt performance prediction in image generation. We also carry out experiments with multiple pre- and post-generation/retrieval performance predictors, providing a set of competitive baselines for future research. We find that a strong supervised pre-generation/retrieval predictor is a worthy competitor for post-generation/retrieval predictors in both text-to-image generation and retrieval. To further demonstrate the usefulness of PQPP, we carry out cross-model, cross-dataset and cross-task experiments, evaluating the generalization capabilities of performance predictors in challenging settings.

In summary, our contribution is threefold:
\begin{itemize}
    \item We propose the first joint benchmark for prompt and query performance prediction.
    \item We collect over 1.6M relevance judgments from human annotators to score a total of 10,200 queries in terms of generation and retrieval performance.
    \item We experiment with multiple pre-generation/retrieval and post-generation/retrieval performance predictors to obtain competitive results, which serve as baselines for future work.
\end{itemize}

\vspace{-0.1cm}
\section{Related Work}
\vspace{-0.1cm}

Studies on QPP initially focused on textual ad hoc retrieval, where both pre- and post-retrieval predictors were considered for sparse retrieval models~\citep{cronen2002predicting,he2004inferring,mothe2005linguistic,yom2005learning}. Some recent studies investigated dense (neural network) retrieval models~\citep{arabzadeh2020neural,datta2022deep,datta2023combining,faggioli2023query,zamani2018neural}, as well as diverse tasks, such as conversational search~\citep{hashemi2019performance,meng2023query,samadi2023performance}. We discuss QPP in textual ad hoc retrieval in Appendix \ref{sec_related_appendix}.   




\noindent
\textbf{QPP in text-to-image retrieval/generation.}
QPP in a multimodal (\eg~text-to-image) context is a relatively new research area~\citep{bizzozzero2023prompt,kumari2023ablating,pavlichenko2023best,valem2023self,Wu-ICCV-2023}. The exploration of QPP in the context of text-to-image retrieval and generation is gaining significant interest, particularly with the rapid advancements in generative methods \citep{Croitoru-TPAMI-2023}. This research domain, distinct for its multimodal nature, aims to enhance the prediction of text query effectiveness for retrieving relevant images. Initial studies, such as those of \citet{xing2010query} and \citet{tian2011query, tian2014query}, have laid the groundwork by exploring query difficulty prediction in image retrieval, utilizing machine learning algorithms and assessing the utility of various features and information sources.

Further contributions by \citet{li2012query} delve into the challenges of estimating query difficulty with unigram language models and visual word verification, highlighting the complexities of aligning text-based queries with visual data. Meanwhile, the development of a self-supervised framework for Content-Based Image Retrieval (CBIR) systems, addressing the scarcity of labeled data through synthetic data and rank-based feature training, marks a significant advancement \citep{valem2023self}.
Recent efforts, such as those of \citet{Wu-ICCV-2023} and \citet{pavlichenko2023best}, focus on integrating human feedback into text-to-image models to refine prediction accuracy and enhance the visual appeal of generated images. \citet{kumari2023ablating} introduced an approach to concept ablation in text-to-image synthesis, aiming to selectively prevent the generation of specific concepts. 

The closest work to our research is that of \citet{bizzozzero2023prompt}, which introduced the concept of \emph{prompt performance prediction}. The authors assessed the effectiveness of prompts in generating images. However, in their study, the ground-truth is automatically generated, which may introduce a significant bias in the evaluation. In contrast, we are the first to explore the prompt performance prediction task with respect to human relevance judgments. In addition, we introduce a benchmark that provides performance measurements for the same set of prompts/queries in both generation and retrieval, enabling the comparative study of QPP across both tasks. Another novel contribution of our work is the study of post-generation prompt performance predictors.

\noindent
\textbf{Human feedback in text-to-image generation.} Human feedback can provide fundamental insights for text-to-image generators, which can harness the provided annotations to increase the quality of the generated images. To this end, several works \cite{Kirstain-NeurIPS-2023,Liang-CVPR-2024,Wu-ICCV-2023,Xu-NeurIPS-2024} collected human feedback for generated images and developed novel benchmarks to train and evaluate generative models. \citet{Kirstain-NeurIPS-2023} obtained preference annotations by asking humans to choose an image from a pair of generated images. \citet{Wu-ICCV-2023} also collected a dataset of human preferences on generated images. The dataset was used to train a classifier to output a human preference score. The classifier was further used to fine-tune Stable Diffusion. \citet{Xu-NeurIPS-2024} asked workers to rank several generated images and rate them based on their quality. The annotations were used to train a reward model, which provided feedback for training diffusion models. Unlike previous studies, \citet{Liang-CVPR-2024} collected rich annotations consisting of fine-grained scores, implausible regions and misaligned words. The annotations were further used to improve image generation.

While we recognize the outstanding merits of previous studies that collected human feedback for  improving text-to-image generation \cite{Kirstain-NeurIPS-2023,Liang-CVPR-2024,Wu-ICCV-2023,Xu-NeurIPS-2024}, we would like to highlight that the focus of our work is different. To our knowledge, our work represents the first attempt to study the task of predicting the performance of text-to-image generators on a given prompt.

\vspace{-0.1cm}
\section{Proposed Benchmark}
\label{sec_data}
\vspace{-0.1cm}

\begin{figure*}[t]
  \centering
  \includegraphics[width=0.99\linewidth]{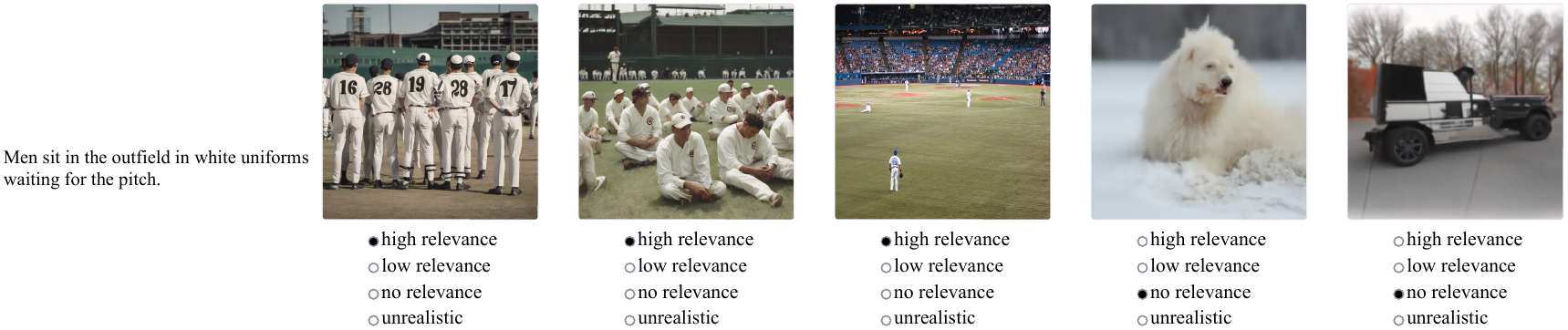}
  \vspace{-0.25cm}
  \caption{An example showing the annotation interface for a random prompt and the images associated with the respective prompt. For each image, the annotator can select one of the following options: high relevance, low relevance, no relevance and unrealistic. The relevance judgments of an annotator are shown for illustrative purposes.  The locations of images are randomly generated, each time they are displayed. Hence, the annotators do not see the images in the same order, which eliminates positional biases. Best viewed in color.}
  \label{fig_anno_interface}
  \vspace{-0.3cm}
\end{figure*}

To develop our novel Prompt and Query Performance Prediction (PQPP) benchmark, we first harness the images and captions from the MS COCO \citep{Lin-ECCV-2014} training set, which comprises approximately 118K images. Each of these images is accompanied by $5$ to $7$ descriptive captions, providing a rich context for our study.

To establish a unified foundation for text-to-image generation and retrieval, we select a subset of $10,000$ captions to be used across tasks either as \emph{queries} for text-to-image retrieval, or as \emph{prompts} for text-to-image generation. We aim for a wide variety of captions that can be clearly associated with images. The selection process entails identifying the most relevant caption for each image (from the small set of captions available in MS COCO for the respective image), where the (caption, image) similarity is measured via the cosine similarity in the CLIP embedding space \citep{Radford-ICML-2021}. This reduces the number of captions to about 118K. The resulting captions are further processed by a sentence transformer based on BERT \citep{Thakur-NAACL-2020} to extract sentence embeddings. Aiming to enhance the distinctiveness of our final set of captions, we next apply the k-means clustering algorithm based on k-means++ initialization on the extracted embeddings. For each cluster, we select the nearest caption to the center of each cluster and include it in our final set of captions. We set $k\!=\!10,\!000$ to obtain 10K prompts/queries. Although MS COCO captions have been used before to train text-to-image generators \cite{Bao-CVPR-2023,Kang-CVPR-2023,Saharia-NeurIPS-2022,Xue-NeurIPS-2024}, prompts written by users can often be more descriptive \cite{Croitoru-CVIU-2024}. To this end, we add all the prompts from DrawBench \cite{Saharia-NeurIPS-2022} to our set, resulting in a combined set of \textbf{10,200} prompts/queries. We next describe the procedures for collecting human relevance judgments on both tasks. 

\vspace{-0.1cm}
\subsection{Prompt Performance Assessment}
\label{sec_prompt_performance_assessment}
\vspace{-0.1cm}

\noindent
\textbf{Generative  models.}
To generate images for our set of prompts, we employ the well-established SDXL \citep{Podell-ICLR-2024} and GLIDE \citep{Nichol-ICML-2021} generative models. For the first model, we choose the Stable Diffusion XL (base-1.0) variant, which produces images with a resolution of $1024\times1024$ pixels. In contrast, GLIDE initially produces images at a resolution of $64\times64$ pixels. The images are subsequently upsampled to a resolution of $256\times256$ pixels using the upsampling model integrated in GLIDE. For each prompt, we generate two images with SDXL and two images with GLIDE. 

\noindent
\textbf{Control and calibration.}
Along with the four generated images, we include the image from MS COCO associated with each prompt into the annotation process. The ground-truth image is included to calibrate user annotations (assuming that the ground-truth image should be labeled as highly relevant) or to exclude annotators that are not seriously engaged in the annotation process. To further ensure the quality of human relevance judgments, we implement a verification process based on a control set of 100 prompts. The control prompts are independently annotated by three authors of this study, following the same protocol as the other annotators. The Fleiss' $\kappa$ coefficient among the control annotators is $0.55$. While running the full-scale annotation process, a control prompt is randomly inserted among every five prompts. This allows us to set a minimum threshold for the Cohen's $\kappa$ coefficient to accept the relevance judgments of an annotator. 
The relevance judgments of an annotator are included into the study if the respective annotator (i) has at least a moderate agreement\footnote{\href{https://en.wikipedia.org/wiki/Cohen\%27s_kappa\#Interpreting_magnitude}{Wiki article on Cohen's $\kappa$: Interpreting Magnitude}} ($\kappa>0.4$) with each annotator in the control group, and (ii) has annotated more than $95\%$ of the ground-truth images as highly relevant.

\noindent
\textbf{Annotation interface.}
We use a custom web interface to collect relevance judgments. The prompts are presented in a random order to each annotator. Each prompt is accompanied by four generated images and one ground-truth image, which are mixed and displayed in a random order, as shown in Figure \ref{fig_anno_interface}. For each image, annotators are given a choice among four evaluative categories: high relevance, low relevance, no relevance, and unrealistic. 
Participants receive thorough instructions (summarized below) before starting the annotation process. The ratings are to be assigned as follows: \emph{high relevance} (score $2$) -- the image depicts over half of the concepts mentioned in the prompt; \emph{low relevance} (score $1$) -- the image captures at least one concept, but fewer than half; \emph{no relevance} (score $0$) -- the image is unrelated, yet realistic; \emph{unrealistic} (score $-1$) -- the image exhibits notable generation artifacts.

\begin{table}[t] 
\centering 
\begin{tabular}{|l|c|c|c|}
 \hline
Statistic & Min & Mean & Max \\
 \hline
 \hline
\#annotations per person & 30 & 1,681 & 15,845\\
Fleiss' $\kappa$ & 0.41 & 0.54 & 1.00 \\
\hline
\end{tabular}
\vspace{-0.25cm}
\caption{Statistics about the annotators enrolled in the annotation process for generated images.} 
\label{tab:annotator_stats} 
\vspace{-0.44cm}
\end{table}

\noindent
\textbf{Annotation process.}
Our aim is to collect at least three annotations per image. In order to collect the required number of annotations (10,200 prompts $\times$ 5 images $\times$ 3 annotations = 153,000 annotations in total), we recruit $173$ annotators. Based on our selection criteria, $26$ annotators are excluded from the process. This leaves us with $147$ valid annotators. Given the asynchronous nature of the annotation process, several prompts ended up having more than three annotations, leading to a total of 247,050 annotations. When a prompt has more than three relevance judgments, we keep the annotations provided by the top three annotators with the highest Cohen's $\kappa$ coefficients (with respect to the control prompts). Some statistics about the enrolled annotators are presented in Table \ref{tab:annotator_stats}. Notably, the Fleiss' $\kappa$ coefficient computed across all annotators is consistent with that of the control annotators. Upon excluding the annotations corresponding to the ground-truth images, we find that most images are voted as highly relevant (see Figure \ref{fig_hist_labels} in Appendix \ref{sec_det_ppa}), confirming that SDXL and GLIDE generally produce relevant results.

\noindent
\textbf{Measuring prompt performance.}
To derive the final prompt performance in image generation, we first map the relevance categories to numerical values, as follows: \emph{high relevance} is mapped to 2, \emph{low relevance} to 1, \emph{no relevance} to 0, and \emph{unrealistic} to -1. 
To exclude outlier annotations, we group the four relevance categories into two high-level categories. The first category, combining the \emph{high relevance} and \emph{low relevance} annotations, represents images that are at least somewhat relevant to the prompt. The second category, combining the \emph{unrealistic} and \emph{no relevance} labels, represents images that are not acceptable for the given prompt, either because they are irrelevant or unrealistic. 
We employ a majority voting mechanism on the high-level categories to decide if a generated image is either \emph{relevant} or \emph{irrelevant}. The Fleiss' $\kappa$ coefficient for these categories is $0.75$, suggesting that the annotations are more consistent at this coarse level. Since there are only two high level categories and three annotations per image, there is no need to break ties (a majority always exists). The majority voting is performed to rule out outlier annotations. The final relevance of an image is given by averaging the scores (between $-1$ and $2$) associated with the votes forming the majority. The performance of a prompt is given by the average relevance score computed across the generated images. We further refer to the resulting score as \emph{human-based prompt performance} (HBPP).

More details about the prompt performance assessment are discussed in Appendix \ref{sec_det_ppa}.

\vspace{-0.1cm}
\subsection{Query Performance Assessment}
\label{sec_query_performance_assessment}
\vspace{-0.1cm}

\noindent
\textbf{Retrieval models.}
For text-to-image retrieval, we employ two distinct vision-language models: CLIP \citep{Radford-ICML-2021} and BLIP-2 \citep{Li-ICML-2023}. For CLIP, we select the ViT-Base \citep{Dosovitskiy2021} architecture with patches of $32\times32$ pixels. For BLIP-2, we choose the ViT-Large backbone. 
These models are pre-trained on natural images, which makes them suitable for image retrieval on MS COCO.



\noindent
\textbf{Annotation process.}
For the retrieval setting, we devise a semi-automatic labeling process to generate reference (ground-truth) relevance judgments for the 10,200 queries. We first employ an automatic process to restrict the number of retrieved images to 2,000 per query. We exploit the structure of the MS COCO dataset based on (image, caption) pairs to generate preliminary relevance judgments using sentence BERT \citep{Thakur-NAACL-2020}. More specifically, we compute the cosine similarity in the embedding space of sentence BERT between each query in PQPP and each caption in MS COCO. Based on a preliminary exploratory data analysis, we set the cosine similarity threshold to $0.7$ to determine a comprehensive set of potentially relevant results. The image corresponding to each caption that has the cosine similarity with a query higher than $0.7$ is added to the preliminary set of relevant results for the respective query. For some short and generic text queries, the preliminary set may contain thousands of potentially relevant images. For such queries, we refine the results ranked below 1,000 using a bag-of-words representation. More specifically, a low-rank image is kept only if the bag-of-words representation of the query is included in the bag-of-words representation of the aggregated captions of the respective image. All queries are limited to 2,000 images in the preliminary set that undergoes manual labeling. 
The preliminary steps described above generate a total of 1,393,363 images, which are further subject to rigorous manual review. There are 100 annotators involved in the manual annotation of the potentially relevant images. The annotators are asked to label each image as relevant or irrelevant to the corresponding query. Each image is annotated by two annotators. An image is kept in the relevant set if it is voted as relevant by one annotator. Two of the evaluators, who are also the main authors of this paper, annotated a set of 4,200 queries. These 4,200 queries are used as control queries for the other enrolled annotators. The remaining queries were randomly divided into batches of 50 queries. In each batch, there are 5 control queries, which are used to exclude annotators that provide poor relevance judgments. There are 98 human evaluators who annotated between 1 and 4 batches. We employed the $F_1$ measure on relevant images for control queries to estimate the quality of the relevance judgments, and set a threshold of $0.4$ to accept annotations. There are 7 annotators who were excluded from the annotation process based on the considered threshold. For the remaining annotators, we obtain a mean $F_1$ score of $0.727$. The minimum $F_1$ score is $0.447$, which is significantly higher than the $F_1$ score of $0.150$ of the random chance baseline. The manual annotation process reduced the total number of relevant images to 530,360. In other words, the annotators removed almost two thirds of the originally retrieved images.

\noindent
\textbf{Measuring query performance.}
To determine the performance level of system for a given query, we employ two alternative measures of retrieval effectiveness, namely 
the precision for the top $n$ retrieved results (P@$n$) and the reciprocal rank (RR). The precision@$n$ is the ratio between the number of true positive images and $n$. Since P@$10$ is often used in text QPP \citep{yom2005learning}, we adopt the same measure and set $n=10$ for our benchmark. The reciprocal rank of a query is given by the ratio between 1 and the rank of the first relevant result. 
We estimate the P@$10$ and RR measures for both CLIP and BLIP-2. 

\vspace{-0.1cm}
\subsection{Evaluation Protocol}
\vspace{-0.1cm}

We divide the annotated prompts/queries into 6,080 for training, 2,040 for validation, and 2,080 for testing. To evaluate performance predictors, we measure the Pearson and Kendall $\tau$ correlation coefficients between the predicted and the ground-truth performance levels of all test queries, following conventional evaluation procedures in text~\citep{yom2005learning,zhao2008effective} and image~\citep{Poesina-SIGIR-2023} QPP. Furthermore, we test the significance of the results with respect to the random chance baseline using Student's t-testing~\citep{Roitman-SIGIR-2018}.

\section{Generation vs.~Retrieval Performance}
\label{Sub:GvsR}

\begin{table}[t]
\centering 
\begin{tabular}{|l|c|c|}
\hline
 Metric & Pearson & Kendall \\
\hline
\hline
 HBPP vs. P@10 & 0.135$^\ddagger$ & 0.093$^\ddagger$ \\ 
 HBPP vs. RR   & 0.072$^\ddagger$ & 0.048$^\dagger$ \\
 P@10 vs. RR   & 0.560$^\ddagger$ & 0.512$^\ddagger$ \\
\hline
\end{tabular}
\vspace{-0.2cm}
\caption{Pearson and Kendall $\tau$ correlation coefficients between the performance levels measured in image generation vs.~image retrieval. According to a Student's t-test, the results marked with $\dagger$ and $\ddagger$ are significantly better than the random chance baseline at p-values ${0.01}$ and ${0.001}$, respectively.}
\vspace{-0.1cm}
\label{tab_gen_vs_ret} 
\end{table}

For each query, our benchmark provides performance measurements in both generation and retrieval settings. Taking advantage of the structure of the PQPP benchmark, we next analyze the correlation between the studied tasks: prompt performance prediction (in text-to-image generation) and query performance prediction (in text-to-image retrieval). We present the correlation results in Table~\ref{tab_gen_vs_ret}. Although the correlations are statistically significant, the empirical analysis reveals surprisingly low correlations between the ground-truth performance measurements for the generative and retrieval tasks. This observation indicates that the tasks are rather orthogonal, confirming that image generation requires the development of dedicated prompt performance predictors.

\begin{figure}[t]
  \centering
  \includegraphics[width=1.0\linewidth]{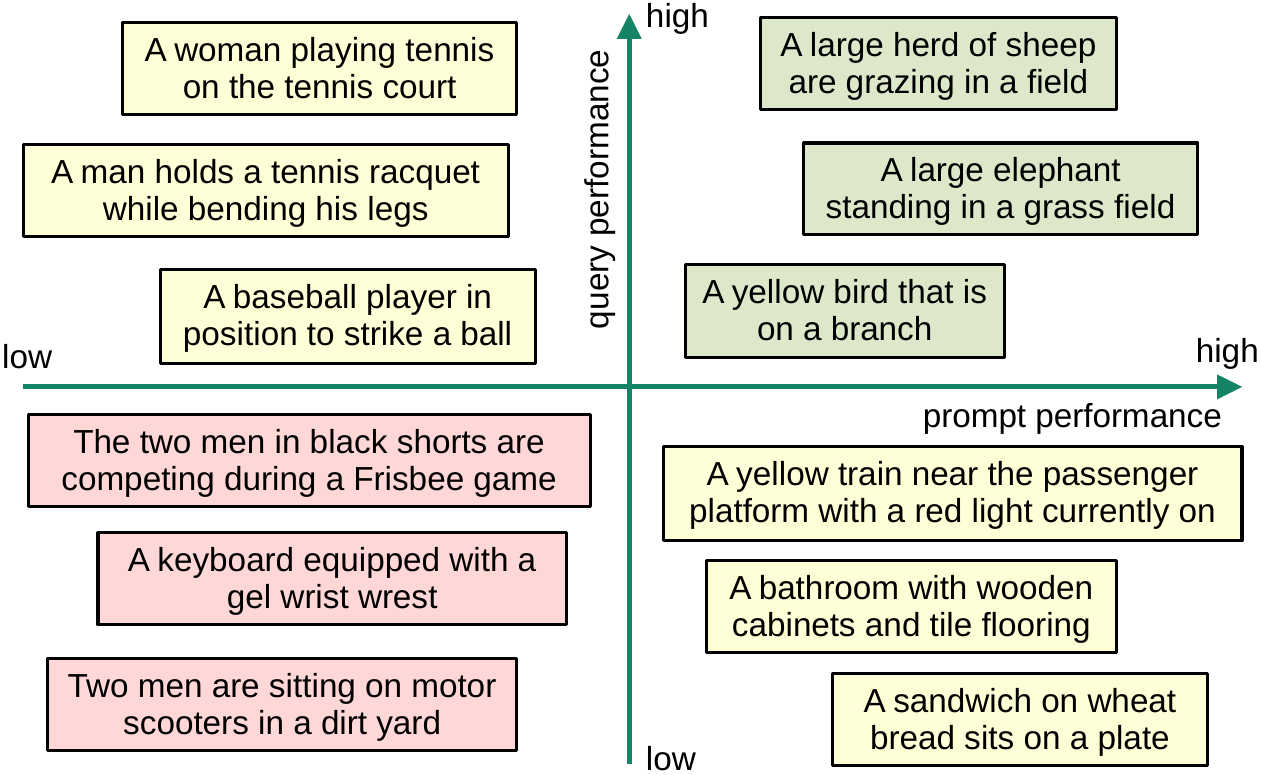}
  \vspace{-0.6cm}
  \caption{Representative prompts/queries that exhibit high or low performance in text-to-image generation (on the horizontal axis) and text-to-image retrieval (on the vertical axis).}
  \label{fig_gen_vs_ret}
  \vspace{-0.3cm}
\end{figure}

Aside from the quantitative results presented in Table~\ref{tab_gen_vs_ret}, we manually inspect prompts/queries from the following four categories: (1) high performance in both generation and retrieval; (2) high performance in generation, low performance in retrieval; (3) low performance in generation, high performance in retrieval; and (4) low performance in both generation and retrieval. We present illustrative captions of the four categories in Figure \ref{fig_gen_vs_ret}. For the first category, the queries usually refer to an animate object performing an action in a location. There are no or few attributes to describe the objects or the location, which increases the likelihood of finding many matches during retrieval, and reduces the constraints during generation. The queries from the second category include inanimate objects with specific descriptions. Generative models seem to handle many specific attributes rather well, but the retrieval models have a hard time finding images that match the specific descriptions. The third category comprises queries about a person performing an action specific to some sport. The MS COCO database contains many images of people playing sports, which makes the retrieval task pretty easy. However, generative models struggle to generate a person in a specific pose, performing a certain action that typically involves interacting with an object, \eg~tennis racket or baseball bat. For the fourth category, representative queries refer to groups of people, \eg~two men, performing specific actions in specific locations, which are rather uncommon. Queries in this category can also refer to uncommon objects with specific attributes, which can even be misspelled, \eg~``gel wrist wrest''. Prompts/queries describing rare situations/objects are difficult for both generative and retrieval models. On the one hand, there is some common ground between prompt/query performance in generation and retrieval, as confirmed by the examples from the first and fourth categories. On the other hand, the examples from the second and third categories indicate that there are task-specific characteristics placing prompt and query performance prediction  at opposite poles. Therefore, we conclude that the task of prompt performance prediction merits further investigation, motivating the utility of our novel benchmark.

\vspace{-0.1cm}
\section{Predictors}
\label{sec_pred}

We briefly present the chosen performance predictors below. We provide details about implementation choices and hyperparameter tuning in Appendix \ref{sec_hid}. 

\vspace{-0.1cm}
\subsection{Pre-generation/retrieval Predictors}
\vspace{-0.1cm}

\noindent
\textbf{Basic text predictors.}
Building on linguistic features to predict query difficulty~\citep{mothe2005linguistic}, we extract a comprehensive suite of pre-retrieval linguistic indicators: the diversity of concepts (total WordNet synsets per prompt/query), lexical density (number of words per prompt/query), morphological complexity (average word length), and the frequency of specific grammatical structures (proper nouns, acronyms, numerals, conjunctions and prepositions). We test a wide variety of basic predictors, but we only report the results of the top two predictors, namely the number of synsets (\#synsets) and the number of words (\#words). Results with the other basic text predictors are reported in Appendix \ref{sec_full_results}.

\noindent
\textbf{Fine-tuned BERT.}
We explore the potential of fine-tuning the BERT model \citep{Devlin-NAACL-2019} as a pre-retrieval performance predictor. We select the \emph{base} architecture based on cased inputs, since the queries contain named entities. The fine-tuning process involves attaching a custom regression head to the pre-trained BERT backbone, which learns to predict the performance of prompts/queries in image generation and retrieval, respectively. 

\vspace{-0.1cm}
\subsection{Post-generation/retrieval Predictors}
\vspace{-0.1cm}

\noindent
\textbf{Fine-tuned CLIP.}
Our first post-generation/retrieval predictor is based on fine-tuning a CLIP model on (query, image) pairs. We use Long-CLIP \cite{Zhang-ECCV-2024} with a ViT-B/32 backbone. We attach a regression head for the image generation scenario, and a binary classification head for the image  retrieval scenario. The utilization of CLIP-based embeddings is aimed at leveraging the model's capacity to jointly represent text and image modalities in a single latent space.
The fine-tuned CLIP learns to predict the relevance judgment of each generated or retrieved image for a given prompt/query. The performance of the prompt/query is then predicted by aggregating the predicted relevance judgments. 

\begin{table*}[t]
\centering
\setlength\tabcolsep{1.8pt}
\small{
\begin{tabular}{|c|l|c|c|c|c|c|c|c|c|c|c|c|c|}
\hline
\multirow{2}{*}{\rotatebox[origin=c]{90}{Predictor Type$\;\;\,$}} & \multirow{6}{*}{Predictor Name} & \multicolumn{4}{c|}{Generative Task} & \multicolumn{8}{c|}{Retrieval Task} \\
\cline{3-14}
  &  & \multicolumn{2}{c|}{GLIDE} & \multicolumn{2}{c|}{SDXL}  &  \multicolumn{4}{c|}{CLIP} & \multicolumn{4}{c|}{BLIP-2} \\
\cline{3-14}
&  & \multicolumn{2}{c|}{HBPP} & \multicolumn{2}{c|}{HBPP}  &  \multicolumn{2}{c|}{P@10} & \multicolumn{2}{c|}{RR} & \multicolumn{2}{c|}{P@10} & \multicolumn{2}{c|}{RR} \\
\cline{3-14}
 & & \rotatebox[origin=c]{90}{$\;$Pearson$\;$} & 
\rotatebox[origin=c]{90}{Kendall} & \rotatebox[origin=c]{90}{Pearson} & \rotatebox[origin=c]{90}{Kendall} & \rotatebox[origin=c]{90}{Pearson} & \rotatebox[origin=c]{90}{Kendall} & \rotatebox[origin=c]{90}{Pearson} & 
\rotatebox[origin=c]{90}{Kendall} & \rotatebox[origin=c]{90}{Pearson} & 
\rotatebox[origin=c]{90}{Kendall} & \rotatebox[origin=c]{90}{Pearson} & 
\rotatebox[origin=c]{90}{Kendall}\\
\hline
\hline
\multirow{3}{*}{\rotatebox[origin=c]{90}{Pre-}} & \#synsets &$-0.112^{\ddagger}$  &$-0.076^{\ddagger}$  &$-0.087^{\ddagger}$ &$-0.080^{\ddagger}$ &$-0.110^{\dagger}$ &$ -0.058^{\ddagger}$  & $-0.034$ & $-0.012$ & $-0.115^{\ddagger}$ & $-0.070^{\ddagger}$ & $-0.038$ & $-0.010$\\
& \#words &
$-0.090^{\dagger}$ & $-0.084^{\ddagger}$ & $-0.105^{\ddagger}$ & $-0.109^{\ddagger}$ & $-0.133^{\ddagger}$ & $-0.104^{\ddagger}$ & $-0.035$ & $-0.026$ & $-0.175^{\ddagger}$ & $-0.136^{\ddagger}$ & $-0.038$ & $-0.015$ \\
 & Fine-tuned BERT & $0.566^{\ddagger}$ & $0.406^{\ddagger}$ & $0.281^{\ddagger}$ & $0.232^{\ddagger}$ & $0.451^{\ddagger}$ & $0.277^{\ddagger}$ & $\mathbf{0.221}^{\ddagger}$ & $\mathbf{0.176}^{\ddagger}$ & $\mathbf{0.511}^{\ddagger}$ & $0.328^{\ddagger}$ & $0.168^{\ddagger}$ & $0.139^{\ddagger}$ \\
\hline
\multirow{2}{*}{\rotatebox[origin=c]{90}{Post-}}   
&Fine-tuned CLIP &  $\mathbf{0.649}^{\ddagger}$ & $\mathbf{0.474}^{\ddagger}$ & $\mathbf{0.380}^{\ddagger}$ & $\mathbf{0.246}^{\ddagger}$ & $\mathbf{0.473}^{\ddagger}$ & $\mathbf{0.299}^{\ddagger}$ & $0.200^{\ddagger}$ & $0.149^{\ddagger}$ & $0.498^{\ddagger}$ & $\mathbf{0.358}^{\ddagger}$ & $0.166^{\ddagger}$ & $0.150^{\ddagger}$ \\
 &Correlation CNN &
$0.548^{\ddagger}$ & $0.393^{\ddagger}$ & $0.159^{\ddagger}$ & $0.107^{\ddagger}$ & $0.270^{\ddagger}$ & $0.186^{\ddagger}$ & $0.189^{\ddagger}$ & $0.162^{\ddagger}$ & $0.159^{\ddagger}$ & $0.133^{\ddagger}$ & $\mathbf{0.206}^{\ddagger}$ & $\mathbf{0.158}^{\ddagger}$ \\
\hline
\end{tabular}
}
\vspace{-0.25cm}
\caption{Results of the prompt/query performance predictors for the generative and retrieval settings on the PQPP test set. On the generative task, we report the correlation of the predicted value with the HBPP performance of SDXL and GLIDE, respectively. On the retrieval task, the correlation is computed for the P@${10}$ and RR scores of CLIP and BLIP-2, respectively. For each task and model, the highest correlation is highlighted in bold. According to a Student's t-test, the results marked with $\dagger$ and $\ddagger$ are significantly better than the random chance baseline at p-values ${0.01}$ and ${0.001}$, respectively. Additional predictors are reported in Appendix \ref{sec_full_results}.}
\label{Tab:results}
\vspace{-0.1cm}
\end{table*}

\begin{table*}[t]
\centering
\setlength\tabcolsep{2.5pt}
\small{%
\begin{tabular}{|c|l|c|c|c|c|c|c|c|c|c|c|c|c|}
\hline
\multirow{2}{*}{\rotatebox[origin=c]{90}{Predictor Type$\;\;\,$}} & \multirow{6}{*}{Predictor Name} & \multicolumn{4}{c|}{Generative Task} & \multicolumn{8}{c|}{Retrieval Task} \\
\cline{3-14}
  &  &  \multicolumn{2}{c|}{SDXL$\rightarrow$GLIDE}  & \multicolumn{2}{c|}{GLIDE$\rightarrow$SDXL} & \multicolumn{4}{c|}{BLIP-2$\rightarrow$CLIP} &  \multicolumn{4}{c|}{CLIP$\rightarrow$BLIP-2} \\
\cline{3-14}
&  & \multicolumn{2}{c|}{HBPP} & \multicolumn{2}{c|}{HBPP}  &  \multicolumn{2}{c|}{P@10} & \multicolumn{2}{c|}{RR} & \multicolumn{2}{c|}{P@10} & \multicolumn{2}{c|}{RR} \\
\cline{3-14}
 & & \rotatebox[origin=c]{90}{$\;$Pearson$\;$} & 
\rotatebox[origin=c]{90}{Kendall} & \rotatebox[origin=c]{90}{Pearson} & \rotatebox[origin=c]{90}{Kendall} & \rotatebox[origin=c]{90}{Pearson} & \rotatebox[origin=c]{90}{Kendall} & \rotatebox[origin=c]{90}{Pearson} & 
\rotatebox[origin=c]{90}{Kendall} & \rotatebox[origin=c]{90}{Pearson} & 
\rotatebox[origin=c]{90}{Kendall} & \rotatebox[origin=c]{90}{Pearson} & 
\rotatebox[origin=c]{90}{Kendall}\\
\hline
\hline
 \multirow{1}{*}{\rotatebox[origin=c]{0}{Pre-}} & Fine-tuned BERT  &
 $0.165^{\ddagger}$ & $0.128^{\ddagger}$ & $0.150^{\ddagger}$ & $0.087^{\dagger}$ & $0.420^{\ddagger}$ & $0.264^{\ddagger}$ & $0.134^{\dagger}$ & $0.112^{\ddagger}$ & $0.431^{\ddagger}$ & $0.270^{\ddagger}$ & $\mathbf{0.165}^{\ddagger}$ & $0.133^{\ddagger}$ \\
\hline
\multirow{2}{*}{\rotatebox[origin=c]{0}{Post-}}   
&Fine-tuned CLIP &
$\mathbf{0.256}^{\ddagger}$ & $\mathbf{0.186}^{\ddagger}$ & $\mathbf{0.179}^{\ddagger}$ & $\mathbf{0.089}^{\dagger}$ & $\mathbf{0.449}^{\ddagger}$ & $\mathbf{0.291}^{\ddagger}$ & $0.121^{\dagger}$ & $0.122^{\ddagger}$ & $\mathbf{0.453}^{\ddagger}$ & $\mathbf{0.337}^{\ddagger}$ & $0.151^{\ddagger}$ & $\mathbf{0.154}^{\ddagger}$ \\

 &Correlation CNN &
$0.131^{\ddagger}$ & $0.078^{\ddagger}$ & $0.096^{\dagger}$ & $0.020$ & $0.155^{\ddagger}$ & $0.130^{\ddagger}$  & $\mathbf{0.206}^{\ddagger}$ & $\mathbf{0.162}^{\ddagger}$ & $0.228^{\ddagger}$ & $0.167^{\ddagger}$ & $0.155^{\ddagger}$ & $0.130^{\ddagger}$ \\
\hline
\end{tabular}
}
\vspace{-0.25cm}
\caption{Cross-model results of the prompt/query performance predictors for the generative and retrieval settings on the PQPP test set. On the generative task, we report the correlation results for two cross-model settings: SDXL$\rightarrow$GLIDE and GLIDE$\rightarrow$SDXL. On the retrieval task, we report the correlations for CLIP$\rightarrow$BLIP-2 and BLIP-2$\rightarrow$CLIP, respectively. For each task, the highest correlation is highlighted in bold. According to a Student's t-test, the results marked with $\dagger$ and $\ddagger$ are significantly better than the random chance baseline at p-values ${0.01}$ and ${0.001}$, respectively.}
\label{Tab:cross_model_results}
\vspace{-0.4cm}
\end{table*}

\noindent
\textbf{Correlation-based CNN.}
Following the work of \citet{sun2018assessing}, we employ a convolutional neural network (CNN) trained on the set of generated or retrieved images, respectively. The CNN model takes a correlation matrix between all image pairs as input. The correlation of an (image, image) pair is given by the cosine similarity between the (pre-trained) CLIP embeddings of the respective images.

\vspace{-0.1cm}
\section{Experiments and Results}
\label{sec_Results}
\vspace{-0.1cm}

\begin{figure}[t]
  \centering
  \includegraphics[width=0.98\linewidth]{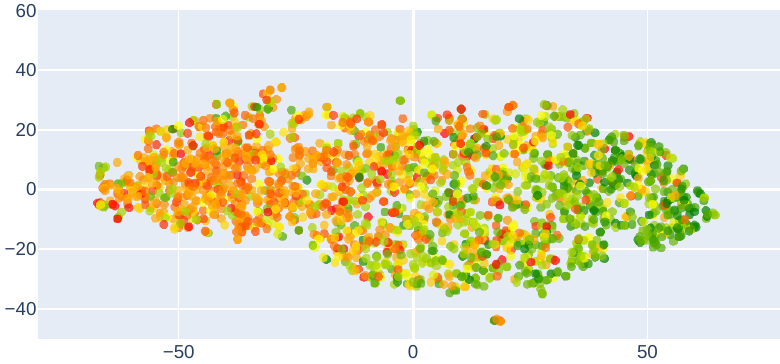}
  \vspace{-0.2cm}
  \caption{t-SNE visualization of the test prompts embedded in the latent space of the BERT predictor fine-tuned on image generation with GLIDE. The ground-truth HBPP performance is encoded via a color map from green (high) to red (low). The visualization confirms that the fine-tuned BERT predictor learns a meaningful representation of the prompts. Best viewed in color.}
  \label{fig_tsne_gen}
  \vspace{-0.4cm}
\end{figure}

\noindent
\textbf{Main results.}
In Table \ref{Tab:results}, we report the results of the best performing predictors for both generative and retrieval tasks (see Table~\ref{Tab:results_full} for additional predictors). For image generation, the fine-tuned CLIP shows the highest correlation with actual performance, achieving a Pearson coefficient of $0.649$ for GLIDE. Interestingly, the second-best predictor is the fine-tuned BERT, which does not even see the generated/retrieved images. While there is a clear ranking of the predictors for the generative task, the correlations reported on the retrieval task do not point towards a clear winner. The top correlations are divided among the fine-tuned BERT, the fine-tuned CLIP and the correlation CNN, respectively. Moreover, we find that it is generally easier to predict P@$10$ than RR. Considering the overall results, an interesting finding of our study is that the pre-retrieval fine-tuned BERT is a worthy competitor for the post-retrieval predictors, being consistently better than the correlation CNN and even surpassing the fine-tuned CLIP in a few cases. In addition, our findings suggest that simple pre-retrieval features, \eg \#synsets or \#words, are not able to capture the complexity of prompts/queries in text-to-image tasks. Nevertheless, statistical testing shows that all the supervised predictors are always significantly better than random chance, indicating that the proposed benchmark is approachable.

Figure \ref{fig_tsne_gen} illustrates a t-SNE visualization of the test prompts embedded in the latent space of the BERT predictor fine-tuned on the generation task. We observe that the learned embedding correlates well with the ground-truth HBPP values, explaining the high accuracy of the fine-tuned BERT predictor on the generation task. An analogous visualization for the retrieval task is shown in Figure \ref{fig_tsne_ret}. 

\noindent
\textbf{Cross-model results.} While the common approach in QPP literature is to predict the performance of a system for a given query, our benchmark enables the assessment of performance predictors across models. With the rapid advancements in neural architectures, novel generative and retrieval models constantly emerge. Therefore, testing the generalization capacity of performance predictors across models is of utter importance. To this end, we report cross-model results on PQPP in Table \ref{Tab:cross_model_results}. We include only supervised predictors in this evaluation, since basic (unsupervised) pre-retrieval predictors do not depend on training data.
On the one hand, we observe that performance predictors exhibit large score drops when tested across generative models, although most of their correlations remain statistically significant. On the other hand, the predictors seem to be more consistent when tested across retrieval models. In a few cases, the correlation scores are even higher when testing is performed across models. For instance, the Pearson correlation of the correlation CNN for the RR measure is $0.189$ when the predictor is trained and tested on CLIP, but the correlation grows to $0.206$ when the training is performed on BLIP-2 and the evaluation is performed on CLIP.

\noindent
\textbf{Additional results.} In Appendix \ref{sec_full_results}, we present the complete set of quantitative experiments, including cross-dataset experiments, cross-task experiments, experiments on individual datasets, and experiments with automatic assessment measures. In Appendix \ref{sec_qual_results}, we analyze the generation/retrieval results from a qualitative perspective. 

\vspace{-0.1cm}
\section{Conclusion and Future Work}
\vspace{-0.1cm}

In this paper, we have presented the first manually-labeled benchmark for prompt performance prediction in the context of prompt-to-image generation. Our benchmark is also applicable for query-to-image retrieval, enabling the direct comparison of the performance prediction task in generation vs.~retrieval scenarios. PQPP is a versatile resource, enabling the evaluation of predictors in various scenarios, such as in-domain, cross-model, cross-dataset and cross-task. Our dataset and code are made publicly available. 

\vspace{-0.05cm}
One direction for future work is to develop a model that combines different pre-retrieval predictors. Combining predictors in a supervised manner has shown its effectiveness in text \citep{dejean2020forward} and image \citep{Poesina-SIGIR-2023} QPP. Another important direction is to organize a shared task associated with the proposed PQPP benchmark, so that the research community can further explore the novel task of prompt performance prediction in image generation.

\noindent
\textbf{Acknowledgments.}
This research is supported by the project ``Romanian Hub for Artificial Intelligence - HRIA'', Smart Growth, Digitization and Financial Instruments Program, 2021-2027, MySMIS no.~334906.

{
    \small
    \bibliographystyle{ieeenat_fullname}
    \bibliography{main}
}

\clearpage
\setcounter{page}{1}
\maketitlesupplementary

\section{Task Usefulness}
\label{sec_usefulness}

In the context of text-to-image generation, if a prompt is predicted as difficult, the system could initiate a conversation to refine the prompt in order to overcome the difficulty and improve the final output. Moreover, the system could indicate to the user its inability to provide a satisfactory image, or it could give positive feedback to the user when a query is predicted as easy. For image generation and image retrieval, when a system is predicted to perform poorly on a prompt/query, additional processes can be activated to improve performance, such as:
\begin{itemize}
    \item Automatic query reformulation: If a query is predicted to perform poorly, it can be automatically reformulated to improve retrieval effectiveness.
    \item Automatic query expansion: For queries expected to perform poorly, QPP can trigger automatic query expansion, adding terms that might improve search performance.
    \item Model selection: Search engines can allocate more computational resources to queries predicted to perform poorly. QPP helps in choosing the most appropriate retrieval algorithm based on the predicted performance for a specific type of query.
    \item Query proposals: Users can be provided with alternative query suggestions if their original query is predicted to perform poorly, improving user satisfaction.
    \item Adapted filtering: In content-based filtering systems, QPP can adapt filtering strategies based on the predicted performance of the query, leading to better results.
\end{itemize}

\begin{figure}[t]
  \centering
  \includegraphics[width=1.0\linewidth]{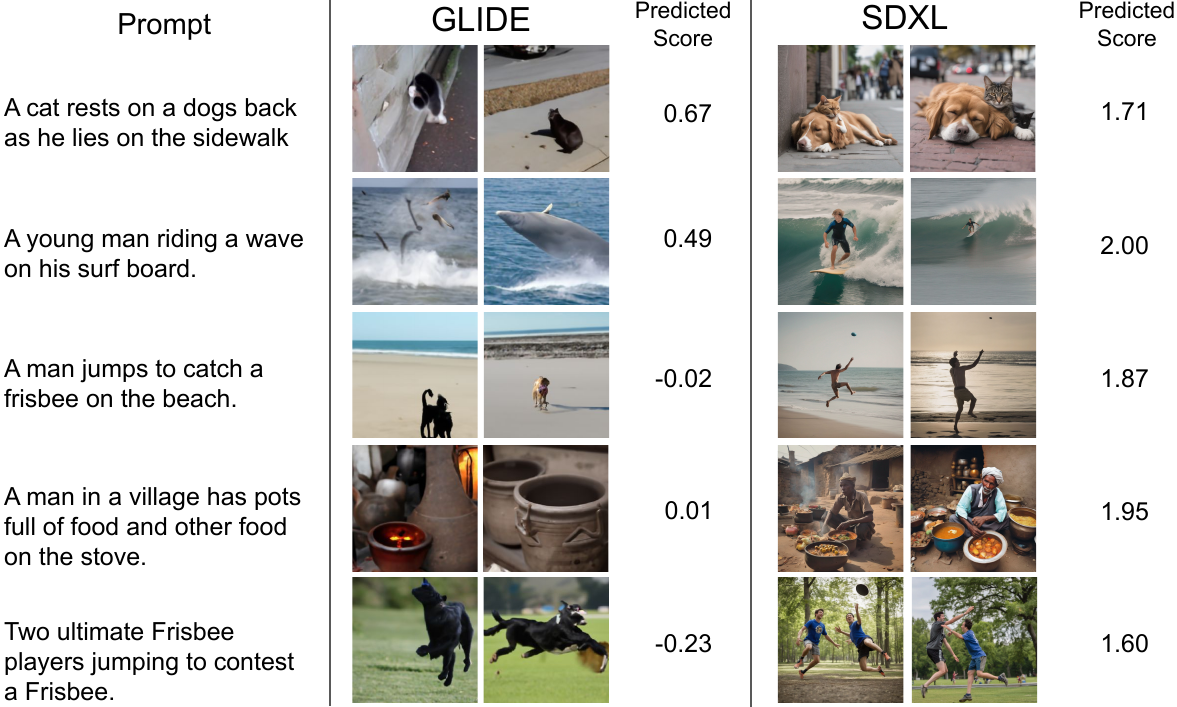}
  \caption{Predicted performance scores by a fine-tuned BERT model for GLIDE and SDXL on various test prompts. The scores are presented alongside the images generated by each model. Best viewed in color.}
  \label{fig_model_sel}
\end{figure}

\begin{figure*}[t]
  \centering
  \includegraphics[width=1.0\linewidth]{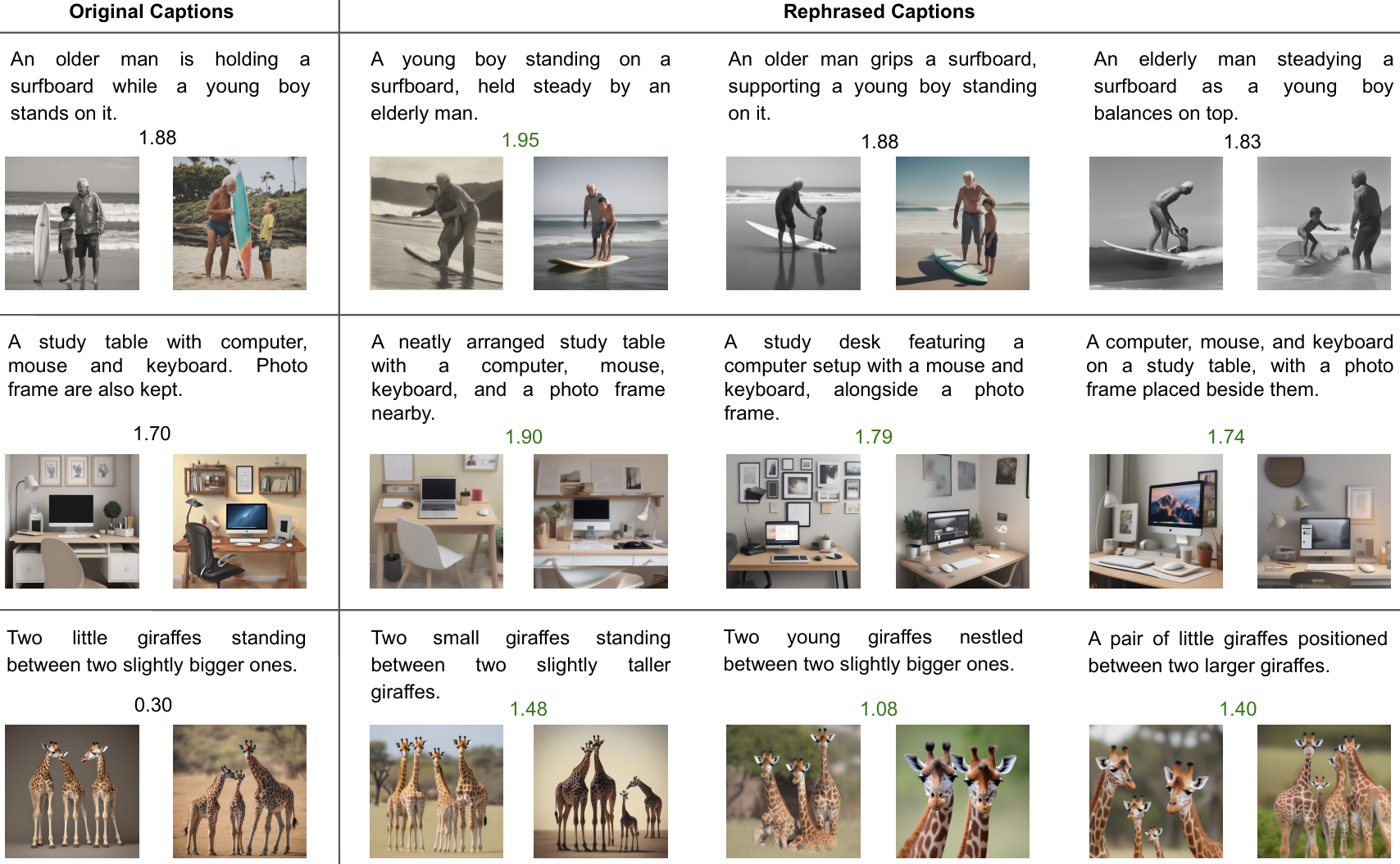}
  \caption{Examples of prompts reformulated by ChatGPT 4.o and associated scores predicted by the fine-tuned BERT performance predictor. The images generated by SDXL for reformulated prompts with higher scores are typically more relevant than those generated for the original prompts. Best viewed in color.}
  \label{fig_query_reform}
\end{figure*}

We further harness the PQPP benchmark and the trained prompt performance predictors to exemplify their utility in two of the applications listed above.

\noindent
\textbf{Use case 1: Generative model selection.}
Next, we illustrate the application in generative model selection via a series of prompt examples that are predicted to perform poorly for GLIDE, but are well-handled by SDXL. Therefore, one could use the more efficient GLIDE model to generate images for easy prompts, and turn to the less efficient (but more capable) SDXL for difficult prompts. The prompt examples, along with their predicted performance, and images generated by GLIDE and SDXL are shown in Figure \ref{fig_model_sel}. The illustrated samples, selected from our test set, indicate that a pre-generation model such as BERT can guide the selection of a more suitable generator, when required.

\noindent
\textbf{Use case 2: Automatic query reformulation.} Another important use case of our benchmark is automatic query reformulation. In Figure \ref{fig_query_reform}, we illustrate examples of rephrased captions by ChatGPT 4.o using the following prompt:
\begin{lcverbatim}
You will need to behave as a large 
language model made to assist with 
query reformulation for the 
application of prompt performance 
prediction. You will need to 
reformulate the query in order to 
increase a text-to-image model to 
its maximum performance. Here are a 
few examples of texts and their 
estimated performance scores:

<Caption>
Black and white of windsurfers on a 
lake.
<Score>
0.15

<Caption>
A black and white picture of several 
buses driving down a street.
<Score>
0.1

<Caption>
A bed made up with linens, is outside 
on a street corner.
<Score>
0.15
\end{lcverbatim}
\begin{lcverbatim}
You will receive a caption and you 
must offer 3 rephrases of the given 
caption, considering the best 
performance for obtaining the desired 
image with a text-to-image model.

<Caption>
[target caption placeholder]
\end{lcverbatim}
Figure \ref{fig_query_reform} shows that the rephrased captions lead to higher scores predicted by the fine-tuned BERT pre-generation predictor, as well as images that are better aligned with the original/rephrased prompt.

\section{Related Work on QPP in Text Retrieval}
\label{sec_related_appendix}

Pre-retrieval features are based on information available prior to the execution of the query. Some are independent to the document collection, such as query length, part-of-speech features (including the number of words of some grammatical categories), query ambiguity~\citep{DeLoupy-LREC-2000}, and query complexity~\citep{mothe2005linguistic}. Other pre-retrieval features depend on the document collection statistics, such as the inverse document frequency~\citep{sparckJones1972}, the query scope~\citep{he2004inferring} (which measures the coverage of a query within the context of a document collection and estimates the proportion of documents that are relevant to the query), and the SCQ~\citep{zhao2008effective} (which is a similarity score between the query and the collection). Pre-retrieval predictors have the huge advantage of being determined before running the search, but they have been found to be less effective than post-retrieval ones on textual ad hoc retrieval~\citep{hauff2008survey,mothe2005linguistic,raiber2014query,Shtok-TIS-2012}.  
Unlike pre-retrieval features, post-retrieval features require conducting document retrieval with the query. Most of these features are calculated based on the scores of the retrieved documents, quantifying the robustness of the document list, or considering the distribution of the document scores~\citep{carmel2010estimating,cronen2002predicting,cummins2011improved,raiber2014query,zhang2018query}. In textual IR, the Clarity Score estimates the specificity of a query considering the language distribution of the document collection and that of the top-retrieved documents~\citep{cronen2002predicting}. The Normalized Query Commitment (NQC), also known as the query drift~\citep{Shtok-TIS-2012}, measures how much the retrieved documents deviate from the central topic of the query. The Weighted Information Gain (WIG) calculates the difference in information content between the documents retrieved for a specific query and a baseline distribution of information in the collection or corpus, based on the scores of the top-retrieved documents~\citep{zhou2007query}.

The main conclusions from the earlier studies on QPP for textual ad hoc retrieval are that post-retrieval predictors outperform pre-retrieval ones~\citep{raiber2014query}, and combinations of predictors using supervised approaches are the most effective~\citep{dejean2020forward}.

Some recent studies investigated QPP on neural IR (NIR) systems~\citep{arabzadeh2020neural,datta2022deep,faggioli2023query,meng2023query,singh2023unsupervised,zamani2018neural}. \citet{datta2022deep} employed convolutional neural layers for their Deep-QPP predictor. This architecture has further been combined with LETOR post-retrieval predictors with some success~\citep{datta2023combining}. According to \citet{faggioli2023query}, QPP models which have been developed for sparse IR methods perform worse when applied to NIR systems. However, the authors did not consider linguistic-based predictors in their work. 
On the other hand, supervised BERT-based QPP models seem to work better. \citet{arabzadeh2021bert} used BERT to predict the performance of search queries in terms of their ability to retrieve relevant documents from a corpus. Such predictors may better capture the semantic aspects of the query-document matching.

Other recent studies focused on the transition from ad hoc search to conversational search~\citep{meng2023query} or question answering~\citep{hashemi2019performance,samadi2023performance}. In conversational search, the experiments showed that supervised QPP methods outperform unsupervised ones when a large amount of training data is available, but unsupervised methods are effective in conversational dense retrieval method assessment. 

\begin{table*}[t]
\centering
\setlength\tabcolsep{1.5pt}
\resizebox{\linewidth}{!}{%
\begin{tabular}{|c|l|c|c|c|c|c|c|c|c|c|c|c|c|}
\hline
\multirow{2}{*}{\rotatebox[origin=c]{90}{Predictor Type$\;\;\,$}} & \multirow{6}{*}{Predictor Name} & \multicolumn{4}{c|}{Generative Task} & \multicolumn{8}{c|}{Retrieval Task} \\
\cline{3-14}
  &  & \multicolumn{2}{c|}{GLIDE} & \multicolumn{2}{c|}{SDXL}  &  \multicolumn{4}{c|}{CLIP} & \multicolumn{4}{c|}{BLIP-2} \\
\cline{3-14}
&  & \multicolumn{2}{c|}{HBPP} & \multicolumn{2}{c|}{HBPP}  &  \multicolumn{2}{c|}{P@10} & \multicolumn{2}{c|}{RR} & \multicolumn{2}{c|}{P@10} & \multicolumn{2}{c|}{RR} \\
\cline{3-14}
 & & \rotatebox[origin=c]{90}{$\;$Pearson$\;$} & 
\rotatebox[origin=c]{90}{Kendall} & \rotatebox[origin=c]{90}{Pearson} & \rotatebox[origin=c]{90}{Kendall} & \rotatebox[origin=c]{90}{Pearson} & \rotatebox[origin=c]{90}{Kendall} & \rotatebox[origin=c]{90}{Pearson} & 
\rotatebox[origin=c]{90}{Kendall} & \rotatebox[origin=c]{90}{Pearson} & 
\rotatebox[origin=c]{90}{Kendall} & \rotatebox[origin=c]{90}{Pearson} & 
\rotatebox[origin=c]{90}{Kendall}\\
\hline
\hline
\multirow{16}{*}{\rotatebox[origin=c]{90}{Pre-}} & \#synsets &$-0.112^{\ddagger}$  &$-0.076^{\ddagger}$  &$-0.087^{\ddagger}$ &$-0.080^{\ddagger}$ &$-0.110^{\dagger}$ &$ -0.058^{\ddagger}$  & $-0.034$ & $-0.012$ & $-0.115^{\ddagger}$ & $-0.070^{\ddagger}$ & $-0.038$ & $-0.010$\\
& \#words &
$-0.090^{\dagger}$ & $-0.084^{\ddagger}$ & $-0.105^{\ddagger}$ & $-0.109^{\ddagger}$ & $-0.133^{\ddagger}$ & $-0.104^{\ddagger}$ & $-0.035$ & $-0.026$ & $-0.175^{\ddagger}$ & $-0.136^{\ddagger}$ & $-0.038$ & $-0.015$ \\
& Average word length  & $0.039$ & $0.041^{\dagger}$ & $-0.067$ & $-0.011$ & $-0.090^{\dagger}$ & $-0.066^{\ddagger}$ & $-0.064^{\dagger}$ & $-0.035$ & $-0.150^{\ddagger}$ & $-0.104^{\ddagger}$ & $-0.116^{\ddagger}$ & $-0.079^{\ddagger}$ \\
& Ratio of proper nouns  & $0.002$ & $-0.027$ & $-0.007$ & $-0.034$ & $-0.053$ & $-0.053^{\dagger}$ & $-0.012$ & $0.001$ & $-0.106^{\ddagger}$ & $-0.102^{\ddagger}$ & $-0. 063^{\dagger}$ & $  -0.040$ \\
& Ratio of acronyms & $0.001$ & $0.008$ &
$0.007$ & $-0.031$& $0.012$ & $-0.000$ & $0.014$ & $0.018$ & $-0.008$ & $-0.028$ & $0.017$ & $-0.002$ \\
& Ratio of numerals & $-0.028$ & $-0.026$ &
$-0.074^{\ddagger}$ & $-0.072^{\ddagger}$ & $-0.049$ & $-0.046$ & $0.007$ & $0.006$ & $-0.065^{\dagger}$ & $-0.070^{\ddagger}$ & $-0.032$ & $-0.025$ \\
& Ratio of conjunctions & $0.054$ & $0.044^{\dagger}$ & $0.037$ & $-0.008$ & $-0.079^{\ddagger}$ & $-0.062^{\ddagger}$ & $-0.024$ & $-0.018$ & $-0.121^{\ddagger}$ & $-0.097^{\ddagger}$ & $-0.032$ & $-0.030$ \\
& Ratio of prepositions & $0.043$ & $0.031$ & $0.033$ & $0.003$  & $0.020$ & $0.020$ & $0.035$ & $0.030$ & $0.014$ & $0.007$ & $0.050$ & $0.038$ \\
& Edge Count & $0.058^{\dagger}$ & $0.084^{\ddagger}$ & $0.020$ & $0.020$ & $0.033$ & $0.047^{\dagger}$ & $0.031$ & $0.011$ & $0.048$ & $0.057$ & $0.030$ & $0.007$ \\
& Edge Weight Sum & $0.054$ & $0.083^{\ddagger}$ & $0.019$ & $0.026$ & $0.033$ & $0.048^{\dagger}$ & $0.030$ & $0.011$ & $0.033$ & $0.057^{\ddagger}$ & $0.028$ & $0.010$ \\
& Inverse Edge Frequency & $0.119^{\ddagger}$ & $0.062^{\ddagger}$ & $0.018$ & $0.039$ & $0.069^{\dagger}$ & $0.039$ & $0.019$ & $0.012$ & $0.046$ & $0.025$ & $0.008$ & $0.011$ \\
& Degree Centrality & $0.073^{\ddagger}$ & $0.071^{\ddagger}$ & $0.022$ & $0.021$ & $0.059^{\dagger}$ & $0.030$ & $0.029$ & $0.010$ & $0.066^{\dagger}$ & $0.038$ & $0.034$ & $0.020$ \\
& Closeness Centrality & $0.032$ & $0.039^{\dagger}$ & $0.133^{\ddagger}$ & $0.048^{\ddagger}$ & $0.077^{\ddagger}$ & $0.035$ & $0.036$ & $0.013$ & $0.048$ & $0.027$ & $0.042$ & $0.010$ \\
& Betweenness Centrality  & $0.025$ & $0.019$ &  $0.062^{\dagger}$ & $0.047^{\dagger}$ & $0.054$ & $0.038$ & $0.026$ & $0.018$ & $0.040$ & $0.035^{\dagger}$ & $0.034$ & $0.027$ \\
& PageRank & $0.064^{\dagger}$ & $0.038$ & $0.088^{\dagger}$ & $0.022$ & $0.022$& $0.021$ & $0.014$ & $0.012$ & $0.049$ & $0.013$ & $0.058^{\dagger}$ & $0.019$ \\

 & Fine-tuned BERT & $0.566^{\ddagger}$ & $0.406^{\ddagger}$ & $0.281^{\ddagger}$ & $0.232^{\ddagger}$ & $0.451^{\ddagger}$ & $0.277^{\ddagger}$ & $\mathbf{0.221}^{\ddagger}$ & $\mathbf{0.176}^{\ddagger}$ & $\mathbf{0.511}^{\ddagger}$ & $0.328^{\ddagger}$ & $0.168^{\ddagger}$ & $0.139^{\ddagger}$ \\
\hline
\multirow{3}{*}{\rotatebox[origin=c]{90}{Post-}}   
&Fine-tuned CLIP &  $\mathbf{0.649}^{\ddagger}$ & $\mathbf{0.474}^{\ddagger}$ & $\mathbf{0.380}^{\ddagger}$ & $\mathbf{0.246}^{\ddagger}$ & $\mathbf{0.473}^{\ddagger}$ & $\mathbf{0.299}^{\ddagger}$ & $0.200^{\ddagger}$ & $0.149^{\ddagger}$ & $0.498^{\ddagger}$ & $\mathbf{0.358}^{\ddagger}$ & $0.166^{\ddagger}$ & $0.150^{\ddagger}$ \\
 &Correlation CNN &
$0.548^{\ddagger}$ & $0.393^{\ddagger}$ & $0.159^{\ddagger}$ & $0.107^{\ddagger}$ & $0.270^{\ddagger}$ & $0.186^{\ddagger}$ & $0.189^{\ddagger}$ & $0.162^{\ddagger}$ & $0.159^{\ddagger}$ & $0.133^{\ddagger}$ & $\mathbf{0.206}^{\ddagger}$ & $\mathbf{0.158}^{\ddagger}$ \\
& HPSv2 & $0.482^{\ddagger}$ & $0.352^{\ddagger}$ & $0.026$ & $0.033$ & - & - & - & - & - & - & - & - \\
\hline
\end{tabular}
}
\vspace{-0.25cm}
\caption{Results of the prompt/query performance predictors for the generative and retrieval settings on the PQPP test set. On the generative task, we report the correlation of the predicted value with the HBPP performance of SDXL and GLIDE, respectively. On the retrieval task, the correlation is computed for the P@${10}$ and RR scores of CLIP and BLIP-2, respectively. For each task and model, the highest correlation is highlighted in bold. According to a Student's t-test, the results marked with $\dagger$ and $\ddagger$ are significantly better than the random chance baseline at p-values ${0.01}$ and ${0.001}$, respectively.}
\label{Tab:results_full}
\vspace{-0.3cm}
\end{table*}

\section{Details on Generated Image Annotation}
\label{sec_det_ppa}

To annotate generated images in terms of relevance, the human annotators are essentially asked to count concepts (objects, attributes, actions) that are both mentioned in the input prompt and present in the generated image. Depending on the number of concepts that are present in the image, the annotators are instructed to label images as follows: high relevance (more than half of the concepts are present), low relevance (less than half of the concepts are present), no relevance (no concept is present), unrealistic (the image contains visible generative artifacts, regardless of the number of concepts). The users are informed that a \emph{concept} can be an object, a property of an object or an activity. For example, the caption ``a white dog catches a Frisbee in its mouth'' contains 5 concepts: the adjective ``white'', the noun ``dog'', the verb ``catch'', the noun ``Frisbee'', and the noun ``mouth''. The users are also given a list of potential generation artifacts: objects with inconsistent appearance (wrong shape, wrong color), counting artifacts (too many / too few object parts of a certain kind), perspective artifacts (different parts of the same object are jointly depicted from visibly different perspectives), structural artifacts (objects have wrong, missing or added parts), etc.

\begin{figure}[t]
  \centering
  \includegraphics[width=1.0\linewidth]{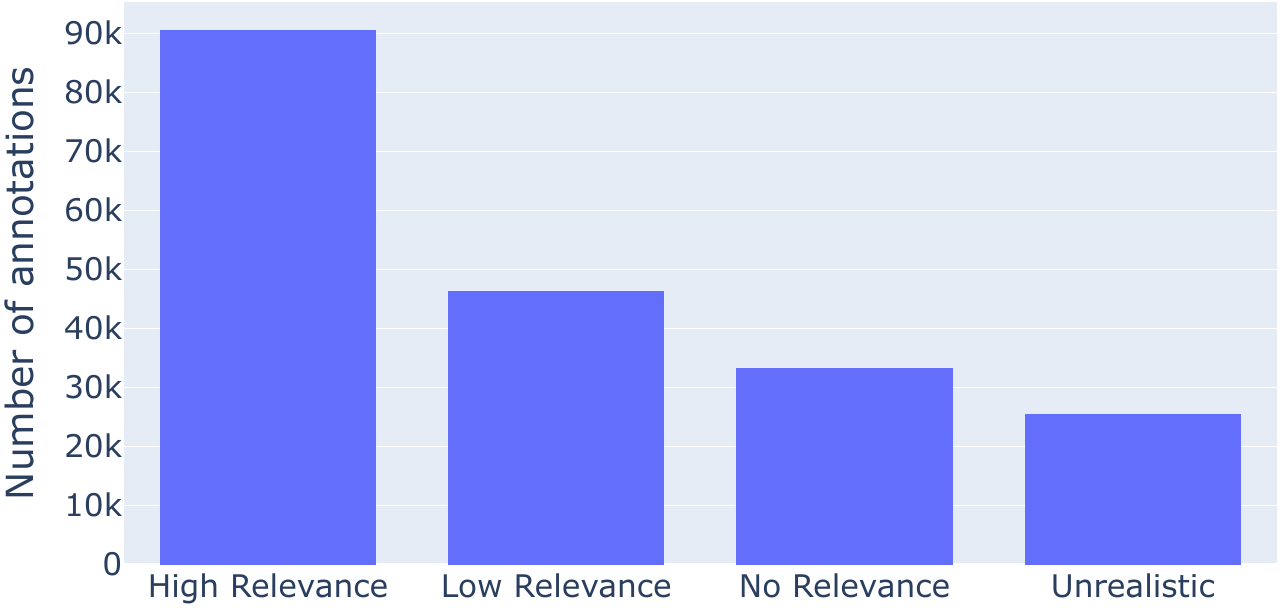}
  \caption{A histogram showing the number of annotations per category for images generated by Stable Diffusion XL and GLIDE.}
  \label{fig_hist_labels}
\end{figure}

In Figure \ref{fig_hist_labels}, we show the number of annotations per category label computed for the four  images generated for each prompt. Although the annotations corresponding to the ground-truth images are excluded, it is clear that most images are voted as highly relevant, confirming that Stable Diffusion and GLIDE generally produce relevant results.

The annotators providing the relevance judgments are adults having at least a bachelor or college degree. The recruited annotators willingly agreed to engage in the annotation process, after reading our terms and conditions. Annotators are allowed to opt out at any time during the annotation process. To reduce bias or uncertainty, annotators are permitted to update previously made annotations or skip specific prompts altogether. Annotators are informed about the inclusion of control prompts within their tasks, but are not given specifics on the frequency of such prompts. A fair compensation (proportional to the number of annotated prompts) is given to each annotator with a Cohen's $\kappa$ coefficient higher than $0.4$ on the control prompts. 

To compute HBPP, we first group the annotations into  relevant (combining \emph{high relevance} and \emph{low relevance} labels) and irrelevant (combining \emph{no relevance} and \emph{unrealistic} labels). We hereby acknowledge that the distinction between \emph{high relevance} and \emph{low relevance} is more difficult to determine, involving a fine assessment of how many of the prompt elements are depicted in the image. This requires evaluators to consider not just the presence of these elements, but also their significance and portrayal within the image, making the distinction between high and low relevance inherently more subjective and challenging. In contrast, the distinction between the high-level categories (\emph{relevant} and \emph{irrelevant}) can be easily assessed.

\section{Predictor Implementation Details}
\label{sec_hid}

\noindent
\textbf{Fine-tuned BERT.}
The regression head consists of a dropout layer and two fully connected layers. The dropout rate is set to $0.3$ to prevent overfitting. The first dense layer is based on ReLU activation functions, and it takes the [CLS] token returned by BERT and transforms it into a $512$-dimensional hidden representation. The second layer contains a single neuron (activated by sigmoid) that predicts prompt/query performance. Before training, the ground-truth performance values are normalized to $[0, 1]$. We employ grid search on the validation set to establish the optimal hyperparameter configuration. More specifically, we consider learning rates between $10^{-3}$ and $10^{-6}$, and weight decays in the set $\{0, 0.1, 0.01\}$. All versions are trained for 15 epochs with early stopping, on mini-batches of $256$ samples. We employ AdamW and optimize the mean squared error (MSE) loss. The fine-tuning is independently carried out for each generation and retrieval model.

\noindent
\textbf{Fine-tuned CLIP.}
For the generative task, the model uses all four images generated by SDXL and GLIDE. For the retrieval task, we limit the training data to the first $25$ images returned by each retrieval model. Although $10$ images would be enough for the P@$10$ metric, estimating the RR measure can require more images. A statistical analysis of the training queries indicates that more than 95\% of the queries have the first relevant image at a rank higher than $25$, which motivates our choice for limiting the training data to $25$ images per model.
The regression/classification head is composed of a two-layer neural network of 512 and 256 neurons, respectively. Both layers are based on ReLU activation. A dropout layer with a drop rate of $0.5$ is added after each dense layer. For the generative task, another layer comprising a single neuron is added to predict prompt performance. The objective of the model is to minimize the MSE loss. For the retrieval task, the last neuron has to determine if an input (query, image) pair is relevant or not. This is a binary classification task, so the model is trained via binary cross-entropy. 
We perform a grid search to find the best hyperparameters, considering learning rates between $10^{-3}$ and $10^{-6}$, and weight decays in the set $\{0, 0.1, 0.01\}$. We employ the AdamW optimizer for $25$ epochs with early stopping, using a batch size of $256$. 

\noindent
\textbf{Correlation-based CNN.}
For the generative task, the size of the input correlation matrix is $4 \times 4$, comprising images generated by both SDXL and GLIDE. For the retrieval task, we apply the same limit to the number of retrieved images per query as for the fine-tuned CLIP predictor. Hence, the size of the correlation matrix for one retrieval model is $25\times 25$. We concatenate the correlation matrices for CLIP and BLIP-2 models in the channel dimension, which results in a tensor of $25\times 25 \times 2$ components that is given as input to the CNN. 

\begin{table*}[t]
\centering
\setlength\tabcolsep{2.3pt}
\small{%
\begin{tabular}{|c|l|c|c|c|c|c|c|c|c|c|c|c|c|}
\hline
\multirow{2}{*}{\rotatebox[origin=c]{90}{Predictor Type$\;\;\,$}} & \multirow{6}{*}{Predictor Name} & \multicolumn{4}{c|}{Generative Task} & \multicolumn{8}{c|}{Retrieval Task} \\
\cline{3-14}
  &  &  \multicolumn{2}{c|}{GLIDE} & \multicolumn{2}{c|}{SDXL}  & \multicolumn{4}{c|}{CLIP} &  \multicolumn{4}{c|}{BLIP-2} \\
\cline{3-14}
&  & \multicolumn{2}{c|}{HBPP} & \multicolumn{2}{c|}{HBPP}  &  \multicolumn{2}{c|}{P@10} & \multicolumn{2}{c|}{RR} & \multicolumn{2}{c|}{P@10} & \multicolumn{2}{c|}{RR} \\
\cline{3-14}
 & & \rotatebox[origin=c]{90}{$\;$Pearson$\;$} & 
\rotatebox[origin=c]{90}{Kendall} & \rotatebox[origin=c]{90}{Pearson} & \rotatebox[origin=c]{90}{Kendall} & \rotatebox[origin=c]{90}{Pearson} & \rotatebox[origin=c]{90}{Kendall} & \rotatebox[origin=c]{90}{Pearson} & 
\rotatebox[origin=c]{90}{Kendall} & \rotatebox[origin=c]{90}{Pearson} & 
\rotatebox[origin=c]{90}{Kendall} & \rotatebox[origin=c]{90}{Pearson} & 
\rotatebox[origin=c]{90}{Kendall}\\
\hline
\hline
 \multirow{1}{*}{\rotatebox[origin=c]{0}{Pre-}} & Fine-tuned BERT
& $0.237^{\dagger}$ & $0.304^{\dagger}$ & $0.167$ & $0.137$ & $-0.064$ & $-0.069$ & $\mathbf{-0.020}$ & ${-0.175} $ & $-0.101$ & $-0.071$ & $-0.063$ & $-0.102$  \\
\hline
\multirow{2}{*}{\rotatebox[origin=c]{0}{Post-}}   
&Fine-tuned CLIP &
$\mathbf{0.417^{\dagger}}$ & $\mathbf{0.317}^{\ddagger}$ & $\mathbf{0.412^{\dagger}}$ & $\mathbf{0.318^{\dagger}}$ & $\mathbf{-0.021}$ & $\mathbf{-0.030}$ & ${-0.198}$ & $-0.019$ & $\mathbf{0.161}$ & $\mathbf{0.129}$ & $0.083$ & $\mathbf{0.138}$ \\

 &Correlation CNN & $0.387^{\dagger}$ & $0.276^{\ddagger}$ & $0.157$ & $0.058$ & $-0.185$ & $-0.098$ & $-0.177$ & $\mathbf{0.117}$ & $0.130$ & $0.092$ & $\mathbf{0.099}$ & $0.018$ \\
\hline
\end{tabular}
}
\vspace{-0.25cm}
\caption{Cross-dataset results of the prompt/query performance predictors for the generative and retrieval settings, using MS COCO for training and DrawBench for testing. On the generative task, we report the correlation of the predicted value with the HBPP performance of SDXL and GLIDE, respectively. On the retrieval task, the correlation is computed for the P@${10}$ and RR scores of CLIP and BLIP-2, respectively. For each task and model, the highest correlation is highlighted in bold. According to a Student's t-test, the results marked with $\dagger$ and $\ddagger$ are significantly better than the random chance baseline at p-values ${0.01}$ and ${0.001}$, respectively.}
\label{Tab:cross_dataset_results_coco_to_draw}
\end{table*}

\begin{table*}[t]
\centering
\setlength\tabcolsep{2.1pt}
\small{%
\begin{tabular}{|c|l|c|c|c|c|c|c|c|c|c|c|c|c|}
\hline
\multirow{2}{*}{\rotatebox[origin=c]{90}{Predictor Type$\;\;\,$}} & \multirow{6}{*}{Predictor Name} & \multicolumn{4}{c|}{Generative Task} & \multicolumn{8}{c|}{Retrieval Task} \\
\cline{3-14}
  &  &  \multicolumn{2}{c|}{GLIDE} & \multicolumn{2}{c|}{SDXL}  & \multicolumn{4}{c|}{CLIP} &  \multicolumn{4}{c|}{BLIP-2} \\
\cline{3-14}
&  & \multicolumn{2}{c|}{HBPP} & \multicolumn{2}{c|}{HBPP}  &  \multicolumn{2}{c|}{P@10} & \multicolumn{2}{c|}{RR} & \multicolumn{2}{c|}{P@10} & \multicolumn{2}{c|}{RR} \\
\cline{3-14}
 & & \rotatebox[origin=c]{90}{$\;$Pearson$\;$} & 
\rotatebox[origin=c]{90}{Kendall} & \rotatebox[origin=c]{90}{Pearson} & \rotatebox[origin=c]{90}{Kendall} & \rotatebox[origin=c]{90}{Pearson} & \rotatebox[origin=c]{90}{Kendall} & \rotatebox[origin=c]{90}{Pearson} & 
\rotatebox[origin=c]{90}{Kendall} & \rotatebox[origin=c]{90}{Pearson} & 
\rotatebox[origin=c]{90}{Kendall} & \rotatebox[origin=c]{90}{Pearson} & 
\rotatebox[origin=c]{90}{Kendall}\\
\hline
\hline
 \multirow{1}{*}{\rotatebox[origin=c]{0}{Pre-}} & Fine-tuned BERT & $0.219^{\ddagger}$ & $0.129^{\ddagger}$ & $0.050$ & $\mathbf{0.086}^{\ddagger}$ & $\mathbf{0.032}$ & $0.028$ & $-0.011$ & $-0.012$ & $-0.026$ & $-0.009$ & $-0.043$ & $-0.034^{\dagger}$  \\
\hline
\multirow{2}{*}{\rotatebox[origin=c]{0}{Post-}}   
&Fine-tuned CLIP & $\mathbf{0.287}^{\ddagger}$ & $\mathbf{0.188}^{\ddagger}$ & $\mathbf{0.078^{\dagger}}$ & $0.034^{\dagger}$ & $0.030$ & $\mathbf{0.040^{\dagger}}$ & $\mathbf{0.086}^{\ddagger}$ & $\mathbf{0.100}^{\ddagger}$ & $-0.041$ & $-0.022$ & $\mathbf{0.070^{\dagger}}$ & $\mathbf{0.054^{\dagger}}$ \\

 &Correlation CNN & $0.194^{\ddagger}$ & $0.130^{\ddagger}$ & $0.047$ & $0.085^{\ddagger}$ &  $-0.094^{\dagger}$ & $-0.081^{\ddagger}$ & $0.045^{\dagger}$ & $0.032^{\dagger}$ & $\mathbf{0.065^{\dagger}}$ & $\mathbf{0.050^{\dagger}}$ & $0.018$ & $0.026$ \\
\hline
\end{tabular}
}
\vspace{-0.25cm}
\caption{Cross-dataset results of the prompt/query performance predictors for the generative and retrieval settings, using DrawBench for training and MS COCO for testing. On the generative task, we report the correlation of the predicted value with the HBPP performance of SDXL and GLIDE, respectively. On the retrieval task, the correlation is computed for the P@${10}$ and RR scores of CLIP and BLIP-2, respectively. For each task and model, the highest correlation is highlighted in bold. According to a Student's t-test, the results marked with $\dagger$ and $\ddagger$ are significantly better than the random chance baseline at p-values ${0.01}$ and ${0.001}$, respectively.}
\label{Tab:cross_dataset_results_draw_to_coco}
\end{table*}

The CNN architecture is composed of four convolutional-pooling blocks, followed by two linear layers. This is a custom architecture that comprises $3\times 3$ convolutional filters applied at a stride of $1$, using a padding of $1$. The number of filters in each of the four convolutional layers is 64, 128, 256 and 512, respectively. A max-pooling is applied after each convolutional layer. The pooling operation uses $2\times 2$ filters applied at a stride of $2$. The first fully connected layer comprises 1024 units. Each hidden neuron is followed by a ReLU activation. The final layer comprises a single neuron that is trained in a regression setting via the MSE loss. The hyperparameter tuning is identical to the one employed for the fine-tuned CLIP model. The correlation-based CNN is trained for $25$ epochs using AdamW with early stopping, on mini-batches of $256$ samples. 

\section{More Quantitative Results}
\label{sec_full_results}

\noindent
\textbf{Results with more predictors.}
In Table \ref{Tab:results_full}, we present the results of all the considered predictors, while Table~\ref{Tab:results} only shows the most interesting ones. We consider that it is important to also report failed attempts with specific predictors. 
The additional predictors are generally based on basic features extracted from queries. The tested predictors are the following: the diversity of concepts (number of WordNet synsets per prompt/query), the lexical density (number of words per prompt/query), the morphological complexity (average word length measured in characters), and the frequency of specific grammatical structures (ratio of proper nouns, ratio of acronyms, ratio of numerals, ratio of conjunctions and ratio of prepositions).
 
Following the work of \citet{Arabzadeh-ECIR-2020}, we implement a suite of predictors based on neural embeddings. In their work, the authors use an ego network to represent each query as a graph. The ego network construction relies on a pre-trained embedding model, such as \emph{word2vec}, which is guided by two hyperparameters: $\alpha$, controlling network depth, and $\beta$, specifying the minimum similarity threshold for node connections.
To build the network, terms directly connected to the root term (ego) must have a similarity of at least $\beta$. For subsequent levels, the similarity threshold is dynamically adjusted as $\beta$ is multiplied with the connecting term's similarity from the previous level. Each child node identifies and connects to its most similar terms meeting this criterion, creating a hierarchical structure.
Graph-based metrics, including {Edge Count}, {Edge Weight Sum}, {Inverse Edge Frequency}, {Degree Centrality}, {Closeness Centrality}, {Betweenness Centrality}, and {PageRank}, are computed over these networks and aggregated to predict query performance.

\begin{table*}[t]
\centering
\setlength\tabcolsep{2.5pt}
\small{%
\begin{tabular}{|c|l|c|c|c|c|c|c|c|c|c|c|c|c|}
\hline
\multirow{2}{*}{\rotatebox[origin=c]{90}{Predictor Type$\;\;\,$}} & \multirow{6}{*}{Predictor Name} & \multicolumn{4}{c|}{Generative Task} & \multicolumn{8}{c|}{Retrieval Task} \\
\cline{3-14}
  &  & \multicolumn{2}{c|}{CLIP$\rightarrow$GLIDE} &  \multicolumn{2}{c|}{BLIP-2$\rightarrow$SDXL}  & \multicolumn{4}{c|}{GLIDE$\rightarrow$CLIP} &  \multicolumn{4}{c|}{SDXL$\rightarrow$BLIP-2} \\
\cline{3-14}
&  & \multicolumn{2}{c|}{P@10$\rightarrow$HBPP} & \multicolumn{2}{c|}{P@10$\rightarrow$HBPP}  &  \multicolumn{2}{c|}{HBPP$\rightarrow$P@10} & \multicolumn{2}{c|}{HBPP$\rightarrow$RR} & \multicolumn{2}{c|}{HBPP$\rightarrow$P@10} & \multicolumn{2}{c|}{HBPP$\rightarrow$RR} \\
\cline{3-14}
 & & \rotatebox[origin=c]{90}{$\;$Pearson$\;$} & 
\rotatebox[origin=c]{90}{Kendall} & \rotatebox[origin=c]{90}{Pearson} & \rotatebox[origin=c]{90}{Kendall} & \rotatebox[origin=c]{90}{Pearson} & \rotatebox[origin=c]{90}{Kendall} & \rotatebox[origin=c]{90}{Pearson} & 
\rotatebox[origin=c]{90}{Kendall} & \rotatebox[origin=c]{90}{Pearson} & 
\rotatebox[origin=c]{90}{Kendall} & \rotatebox[origin=c]{90}{Pearson} & 
\rotatebox[origin=c]{90}{Kendall}\\
\hline
\hline
 \multirow{1}{*}{\rotatebox[origin=c]{0}{Pre-}} & Fine-tuned BERT&  $\mathbf{0.108}^{\ddagger}$ & $\mathbf{0.071}^{\ddagger}$ & $\mathbf{0.103}^{\ddagger}$ & $\mathbf{0.157}^{\ddagger}$ & $\mathbf{0.165}^{\ddagger}$ & $\mathbf{0.174}^{\ddagger}$ & $\mathbf{0.118}^{\ddagger}$ & $\mathbf{0.109}^{\ddagger}$ & $0.155^{\ddagger}$ & $\mathbf{0.167}^{\ddagger}$ & $0.094^{\dagger}$ & $0.087^{\ddagger}$ \\
\hline
\multirow{2}{*}{\rotatebox[origin=c]{0}{Post-}}   
&Fine-tuned CLIP &
$0.075^{\dagger}$ & $0.039$ & $0.092^{\dagger}$ & $0.121^{\ddagger}$ & $0.134^{\ddagger}$ & $0.103^{\ddagger}$ & $0.090^{\dagger}$ & $0.071^{\ddagger}$ & $\mathbf{0.174}^{\ddagger}$ & $0.155^{\ddagger}$ & $\mathbf{0.135}^{\ddagger}$ & $\mathbf{0.114}^{\ddagger}$ \\

 &Correlation CNN &
$0.111^{\ddagger}$ & $0.066^{\dagger}$ & $0.080^{\ddagger}$ & $0.037$ & $0.053$ & $0.037$ & $0.030$ & $0.024$ & $0.026$ & $0.022$ & $0.030$ & $0.022$ \\
\hline
\end{tabular}
}
\vspace{-0.25cm}
\caption{Cross-task results of the prompt/query performance predictors on the PQPP benchmark. We report the correlation results for two cross-task model pairs: (GLIDE, CLIP) and (SDXL, BLIP-2). This pairing generates the following evaluation cases: CLIP$\rightarrow$GLIDE, BLIP-2$\rightarrow$SDXL, GLIDE$\rightarrow$CLIP and SDXL$\rightarrow$BLIP-2. For each case, the highest correlation is highlighted in bold. According to a Student's t-test, the results marked with $\dagger$ and $\ddagger$ are significantly better than the random chance baseline at p-values ${0.01}$ and ${0.001}$, respectively.}
\label{Tab:cross_task_results}
\end{table*}

\begin{table*}[t]
\centering
\setlength\tabcolsep{1.8pt}
\resizebox{0.995\linewidth}{!}{%
\begin{tabular}{|c|c|l|c|c|c|c|c|c|c|c|c|c|c|c|}
\hline
\multirow{6}{*}{\rotatebox[origin=c]{00}{Dataset}} & \multirow{2}{*}{\rotatebox[origin=c]{90}{Predictor Type$\;\;\,$}} & \multirow{6}{*}{Predictor Name} & \multicolumn{4}{c|}{Generative Task} & \multicolumn{8}{c|}{Retrieval Task} \\
\cline{4-15}
 & &  &  \multicolumn{2}{c|}{GLIDE} & \multicolumn{2}{c|}{SDXL}  & \multicolumn{4}{c|}{CLIP} &  \multicolumn{4}{c|}{BLIP-2} \\
\cline{4-15}
& &  & \multicolumn{2}{c|}{HBPP} & \multicolumn{2}{c|}{HBPP}  &  \multicolumn{2}{c|}{P@10} & \multicolumn{2}{c|}{RR} & \multicolumn{2}{c|}{P@10} & \multicolumn{2}{c|}{RR} \\
\cline{4-15}
& & & \rotatebox[origin=c]{90}{$\;$Pearson$\;$} & 
\rotatebox[origin=c]{90}{Kendall} & \rotatebox[origin=c]{90}{Pearson} & \rotatebox[origin=c]{90}{Kendall} & \rotatebox[origin=c]{90}{Pearson} & \rotatebox[origin=c]{90}{Kendall} & \rotatebox[origin=c]{90}{Pearson} & 
\rotatebox[origin=c]{90}{Kendall} & \rotatebox[origin=c]{90}{Pearson} & 
\rotatebox[origin=c]{90}{Kendall} & \rotatebox[origin=c]{90}{Pearson} & 
\rotatebox[origin=c]{90}{Kendall}\\
\hline
\hline
\multirow{2}{*}{\rotatebox[origin=c]{0}{MS COCO}} & \multirow{1}{*}{\rotatebox[origin=c]{0}{Pre-}} & Fine-tuned BERT 
&$0.550^{\ddagger}$  &$0.400^{\ddagger}$  &$0.254^{\ddagger}$ &$0.244^{\ddagger}$ 
&$0.454^{\ddagger}$ &$ 0.271^{\ddagger}$  & $0.257^{\ddagger}$ & $0.197^{\ddagger}$ 
& $0.489^{\ddagger}$ & $0.320^{\ddagger}$ & $0.149^{\ddagger}$ & $0.112^{\ddagger}$\\

 & \multirow{1}{*}{\rotatebox[origin=c]{0}{Post-}}  & Fine-tuned CLIP 
 & $0.657^{\ddagger}$ & $0.479^{\ddagger}$ & $0.360^{\ddagger}$ & $0.245^{\ddagger}$ 
 & $0.435^{\ddagger}$ & $0.315^{\ddagger}$ & ${0.127}^{\ddagger}$ & ${0.105}^{\ddagger}$ 
 & ${0.488}^{\ddagger}$ & $0.399^{\ddagger}$ & $0.058$ & $0.097^{\ddagger}$ \\
\hline
\multirow{2}{*}{\rotatebox[origin=c]{0}{DrawBench}} & \multirow{1}{*}{\rotatebox[origin=c]{0}{Pre-}}   
& Fine-tuned BERT 
&  ${0.358}^{\ddagger}$ & ${0.274}^{\ddagger}$ & ${0.511}^{\ddagger}$ & ${0.216}^{\dagger}$ 
& ${-0.050}$ & ${-0.046}$ & $-0.152$ & $-0.149$ 
& $-0.162$ & ${-0.102}$ & $0.016$ & $0.024$ \\

& \multirow{1}{*}{\rotatebox[origin=c]{0}{Post-}} & Fine-tuned CLIP 
& $0.456^{\ddagger}$ & $0.335^{\ddagger}$ & $0.462^{\ddagger}$ & $0.205$ 
& $-0.118$ & $-0.109$ & $-0.161$ & $-0.128$ 
& $-0.060$ & $-0.039$ & ${-0.048}$ & ${-0.348}^{\dagger}$ \\
\hline
\end{tabular}
}
\vspace{-0.2cm}
\caption{Results of prompt/query performance predictors on MS COCO vs.~DrawBench. On the generative task, we report the correlation of the predicted value with the HBPP performance of SDXL and GLIDE, respectively. On the retrieval task, the correlation is computed for the P@${10}$ and RR scores of CLIP and BLIP-2, respectively. According to a Student's t-test, the results marked with $\dagger$ and $\ddagger$ are significantly better than the random chance baseline at p-values ${0.01}$ and ${0.001}$, respectively.}
\label{Tab:results_coco_vs_draw}
\end{table*}

In general, we find that predictors based on simple heuristics are not capable of capturing prompt/query performance, showcasing typically low correlations, under $0.1$. The predictors based on ego networks \cite{Arabzadeh-ECIR-2020} do not seem to be any better. We perform an additional experiment with the pre-trained HPSv2 \cite{Wu-arXiv-2023} model, employing it to predict the HBPP scores. This model is not as good as the fine-tuned predictors, failing to predict HBPP for SDXL. In general, we find that the only predictors able to consistently predict performance across all models and tasks are the supervised ones, namely the fine-tuned BERT, the fine-tuned CLIP and the correlation CNN.

\noindent
\textbf{Cross-dataset results.}
In Table \ref{Tab:cross_dataset_results_coco_to_draw}, we present results of supervised predictors trained on prompts/queries from MS COCO and tested on prompts/queries from DrawBench. Conversely, in Table \ref{Tab:cross_dataset_results_draw_to_coco}, we show the results of the same predictors trained on DrawBench and evaluated on MS COCO. We first observe that the cross-dataset results are generally higher for the image generation task than for the image retrieval task. This observation can be attributed to the fact that many of the DrawBench queries (around 50\%) have no relevant results in the MS COCO database (as per the collected ground-truth annotations), which places the respective queries in the ``very difficult'' zone. This exacerbates the distribution gap between MS COCO and DrawBench in the retrieval setting. Therefore, it is very challenging for predictors to generalize across datasets. 
Comparing the two scenarios, MS COCO$\rightarrow$DrawBench vs.~DrawBench$\rightarrow$MS COCO, in the image generation context, we find that training on MS COCO leads to better results. This can be attributed to the fact that the number of prompts from MS COCO (10K) is much higher than the number of prompts from DrawBench (200), even after applying our filtering based on k-means to select captions from MS COCO. Nevertheless, both cross-dataset settings are difficult, opening a new avenue for future research: proposing prompt/query performance predictors able to generalize across different data distributions.

\begin{table*}[]
    \centering
    \setlength\tabcolsep{1.8pt}
\begin{tabular}{|c|l|c|c|c|c|c|c|c|c|c|c|c|c|}
\hline
\multirow{2}{*}{\rotatebox[origin=c]{90}{Predictor Type$\;\;\,$}} & \multirow{6}{*}{Predictor Name} & \multicolumn{4}{c|}{Generative Task} & \multicolumn{8}{c|}{Retrieval Task} \\
\cline{3-14}
  &  & \multicolumn{2}{c|}{GLIDE} & \multicolumn{2}{c|}{SDXL}  &  \multicolumn{4}{c|}{CLIP} & \multicolumn{4}{c|}{BLIP-2} \\
\cline{3-14}
&  & \multicolumn{2}{c|}{HPSv2} & \multicolumn{2}{c|}{HPSv2}  &  \multicolumn{2}{c|}{CLIP-P@10} & \multicolumn{2}{c|}{CLIP-RR} & \multicolumn{2}{c|}{CLIP-P@10} & \multicolumn{2}{c|}{CLIP-RR} \\
\cline{3-14}
 & & \rotatebox[origin=c]{90}{$\;$Pearson$\;$} & 
\rotatebox[origin=c]{90}{Kendall} & \rotatebox[origin=c]{90}{Pearson} & \rotatebox[origin=c]{90}{Kendall} & \rotatebox[origin=c]{90}{Pearson} & \rotatebox[origin=c]{90}{Kendall} & \rotatebox[origin=c]{90}{Pearson} & 
\rotatebox[origin=c]{90}{Kendall} & \rotatebox[origin=c]{90}{Pearson} & 
\rotatebox[origin=c]{90}{Kendall} & \rotatebox[origin=c]{90}{Pearson} & 
\rotatebox[origin=c]{90}{Kendall}\\
\hline
\hline
 \multirow{1}{*}{\rotatebox[origin=c]{0}{Pre-}} & Fine-tuned BERT 
&$0.806^{\ddagger}$  &$0.608^{\ddagger}$  &$0.696^{\ddagger}$ &$0.505^{\ddagger}$ 
&$0.437{\ddagger}$ &$ 0.255{\ddagger}$  & $0.207{\ddagger}$ & $ 0.167{\ddagger}$ 
& $0.495{\ddagger}$ & $0.329{\ddagger}$ & $0.144{\ddagger}$ & $ 0.110{\ddagger}$\\

\multirow{1}{*}{\rotatebox[origin=c]{0}{Post-}}  & Fine-tuned CLIP 
 & $0.257^{\ddagger}$ & $0.169^{\ddagger}$ & $0.729^{\ddagger}$ & $0.530^{\ddagger}$ 
 & $0.463^{\ddagger}$ & $ 0.305^{\ddagger}$ & ${ 0.160}^{\ddagger}$ & ${0.122}^{\ddagger}$ 
 & ${0.484}^{\ddagger}$ & $0.358^{\ddagger}$ & $0.159^{\ddagger}$ & $0.144^{\ddagger}$ \\
\hline
\end{tabular}
\vspace{-0.2cm}
\caption{Results of performance predictors for automatic relevance judgments. On the generative task, we report the correlation of the predicted value with the HPSv2 performance of SDXL and GLIDE, respectively. On the retrieval task, the correlation is computed for the CLIP-based P@${10}$ and CLIP-based RR scores of CLIP and BLIP-2, respectively. According to a Student's t-test, the results marked with $\dagger$ and $\ddagger$ are significantly better than the random chance baseline at p-values ${0.01}$ and ${0.001}$, respectively.}
\label{Tab:results_automatic}
\end{table*}

\noindent
\textbf{Cross-task results.}
Although the correlations between the ground-truth scores for image generation and image retrieval are moderate (see Table \ref{tab_gen_vs_ret}), we also aim to assess how well predictors perform across tasks. To this end, we present cross-task results for two model pairs, namely (GLIDE, CLIP) and (SDXL, BLIP-2), in Table \ref{Tab:cross_task_results}. As expected, the correlation coefficients are typically low, indicating that predictors are not able to generalize across tasks. However, this apparent inability of the predictors should be attributed to the low correlations between the image generation and retrieval tasks reported in Table \ref{tab_gen_vs_ret}, which clearly indicate that the two tasks are not very well aligned. 

\noindent
\textbf{MS COCO vs.~DrawBench.}
To assess the disparity between MS COCO and DrawBench, we train and test the fine-tuned BERT and fine-tuned CLIP predictors on the individual subsets (see Table \ref{Tab:results_coco_vs_draw}). On the generative task, predictors obtain comparable results across the two datasets. Since DrawBench is specifically designed for text-to-image generation, its queries are too difficult for the retrieval setup, so predictors fail in this case. In contrast, MS COCO queries have about the same difficulty (on average) in generation and retrieval. This supports our decision to include more captions from MS COCO than DrawBench into PQPP.

\noindent
\textbf{Results for automatic metrics.}
We conduct additional experiments with automatic evaluation metrics instead of the proposed metrics based on human relevance judgments. More specifically, we rely on HPSv2 \cite{Wu-arXiv-2023} for generated images and CLIP for retrieved images. We report the corresponding results in Table \ref{Tab:results_automatic}. Predictors seem to have higher correlation with HPSv2 than with human labels (in image generation), indicating that automatic labels are easier to predict.

\section{More Qualitative Results}
\label{sec_qual_results}

\begin{figure}[t]
  \centering
  \includegraphics[width=1.0\linewidth]{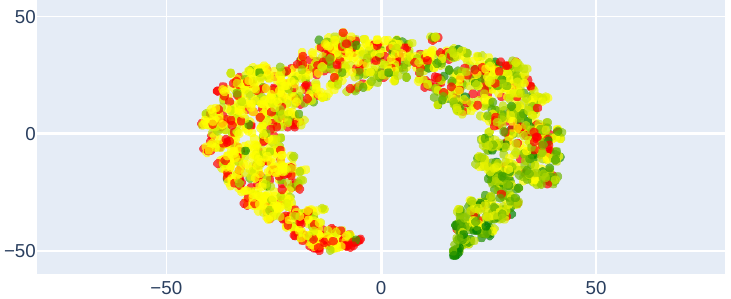}
  \vspace{-0.6cm}
  \caption{t-SNE visualization of the test queries embedded in the latent space of the fine-tuned BERT on image retrieval with BLIP-2. The ground-truth P@${10}$ performance is encoded via a color map from green (high) to red (low). The visualization confirms that the fine-tuned BERT predictor learns a meaningful representation of the queries. Best viewed in color.}
  \label{fig_tsne_ret}
\end{figure}

In Figure \ref{fig_tsne_ret}, we present a t-SNE visualization of the test queries embedded in the latent space of the BERT predictor fine-tuned on image retrieval with BLIP-2. We observe that the learned latent space correlates well with the ground-truth P@$10$ values, explaining the high accuracy of the fine-tuned BERT predictor on the retrieval task. The separation between easy and difficult queries is evident in the retrieval setting, which is consistent with the quantitative results reported in Table \ref{Tab:results}, where the fine-tuned BERT exhibits generally higher Pearson and Kendall $\tau$ correlation coefficients than other predictors.

\begin{figure}[tbh]
  \centering
  \includegraphics[width=1.0\linewidth]{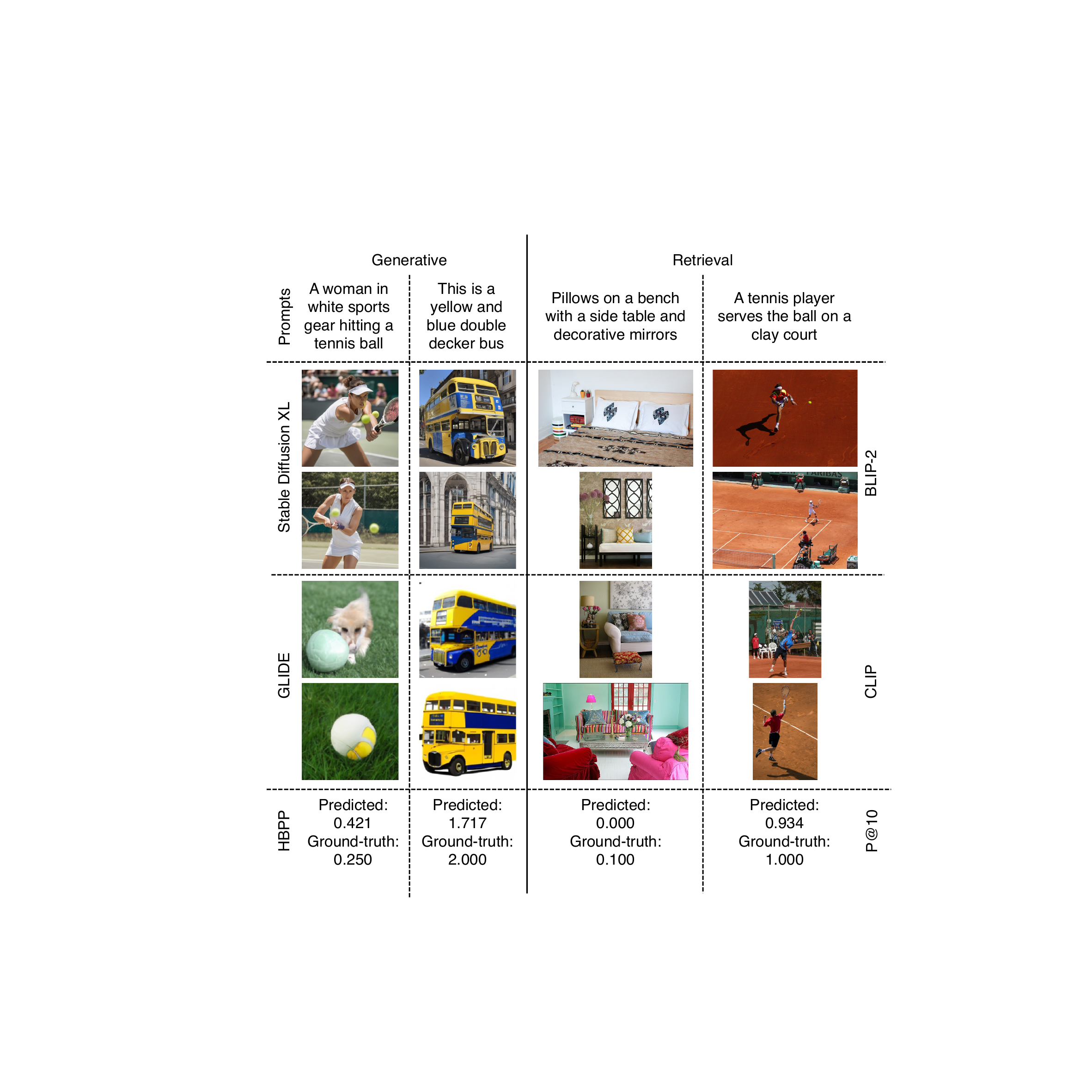}
  \caption{Examples of generative (first two columns) and retrieval (last two columns) results for difficult (first and third columns) and easy (second and fourth columns) queries. For the retrieval systems, we show only the top two results. The prompt/query performance values, namely HBPP and P@${10}$, are predicted by the correlation-based CNN. Best viewed in color.}
  \label{fig_examples}
\end{figure}

We showcase examples of easy and difficult prompts/queries for the generation and retrieval tasks in Figure \ref{fig_examples}.
The generative models exhibit a clear proficiency with prompts referring to inanimate objects, generating images with high relevance. However, their capability falls short when faced with more intricate prompts involving complex actions based on human-object interactions, leading to inaccuracies in object composition. Such cases exhibit artifacts, such as duplicate or missing body parts and misplaced objects, showcasing the lack of deep understanding in both generative models. 
The retrieval systems are capable of fetching images for prompts centered around single, loosely-defined objects. However, they struggle when the prompts require images containing multiple, specific elements, retrieving results that only partially match the query. This limitation highlights a gap in the ability of retrieval systems to interpret and respond to the multifaceted nature of some queries.
This exploration into both generative and retrieval tasks underscores the nuanced challenges faced by systems in accurately capturing and responding to the inherent complexity of certain prompts. It also reinforces the importance of the prompt/query performance prediction task, setting realistic expectations for the outcomes of both generative and retrieval models, based on the detailed content and structure of the prompts/queries.

\section{Limitations}
\label{sec_limitations}

We recognize specific limitations in the annotation processes for both generative and retrieval tasks. In the generative setting, prompts can be subjectively interpreted by users, which may introduce variability in the results. To mitigate this, we incorporated a control set, excluded annotations from annotators who failed on the control set, and computed an average score from multiple user annotations for each image/prompt pair, enhancing the robustness of the evaluations.

In the retrieval setting, the initial ground-truth image bank was generated using a method that combines Sentence-BERT and the bag-of-words model. We acknowledge that some images may contain content that is not fully or accurately captured by their paired captions, which could introduce occasional false negatives or false positives in the ground-truth collection. To minimize the number of false negatives, we set a relatively low similarity threshold for the inclusion of candidate images. Then, the false positives were curated by the enrolled annotators.


\section{Potential Negative Societal Impact}
\label{sec_negative_impact}

The development and deployment of text-to-image generation and retrieval systems come with several societal implications that warrant careful consideration. Here, we outline key areas of concern. Our enhanced dataset is built upon pre-existing datasets, which may inherit and perpetuate biases present in the original data. Additionally, user annotations might have been influenced by their own cultural backgrounds, potentially introducing subjective biases into the final decisions. The pre-trained generative models employed in our study could also exhibit inherent biases, affecting the generated outputs. These combined factors could lead to unfair representations or reinforce existing stereotypes. We acknowledge the necessity of ongoing efforts to identify, measure, and mitigate these biases to ensure the fairness and inclusiveness of our models. In our work, we address these concerns by promoting transparency in our methodology and unifying multiple user annotations to mitigate possible individual biases.

The computational resources required for data collection, filtering, and model training contribute to energy consumption and carbon emissions. We recognize the environmental impact of our work and emphasize the importance of optimizing computational processes and exploring sustainable practices to reduce the ecological footprint of AI research.  

Moreover, our benchmark can be used to develop and improve generative models. Such models can further be used in unethical scenarios, \eg~to generate deep fakes. In recent years, an increase in deep fake materials flooded the web, either to spread false information or to steal sensitive information by impersonating trustworthy individuals. While we strongly believe in the benefits of very capable generative models, we are aware of the potential risks. However, we can see that governments are working very closely with academia and industry on safely developing artificial intelligence, and thus observe and support the increasing focus on models that detect AI-generated content to mitigate the aforementioned risks. 

\section{Ethical Considerations on Data Annotation}
\label{sec_annotation_ethics}

Data annotation by students is a common practice in our host institutions 
and we followed the standard protocols to get  approvals from the corresponding ethics committees. The enrolled students were compensated with bonus points. We would like to emphasize that the students understood that the annotation task is optional, and they could also get the extra bonus points by performing alternative tasks (which did not involve data annotation). Moreover, all students were given the opportunity to obtain a full grade without the optional annotation task. Hence, there was no obligation for any of the students to perform the annotations. The students were also able to opt out, at any time during the annotation, without any penalties. 


\section{Computational Resources}
\label{sec_resources}

We have employed two types of machines to perform our experiments:
\begin{itemize}
\item \textbf{Local Hardware:}
\begin{itemize}
    \item \textbf{GPU:} NVIDIA RTX 3090 with 24GB VRAM
    \item \textbf{CPU:} Intel i9-10920X @ 3.50GHz
    \item \textbf{Memory:} 64GB RAM
    \item \textbf{Storage:} 1TB SSD, 5TB HDD
\end{itemize}
\item \textbf{Cloud VM:}
\begin{itemize}
    \item \textbf{GPU:} NVIDIA A100 with 40GB VRAM
    \item \textbf{CPU:} 12 vCPUs
    \item \textbf{Memory:} 85GB RAM
    \item \textbf{Storage:} 100GB HDD
\end{itemize}
\end{itemize}

Our annotation platform was hosted using Google Cloud Provider, with authentication developed with Google Firebase Authentication, and image hosting facilitated by Google Cloud Storage. By detailing the utilized computational resources, we aim to provide transparency and reproducibility for our research.

\section{Computational Time Estimation}
\label{sec_computational_time}

We present the following estimation of the compute time (in hours) required to fully replicate the experiments detailed in this paper:
\begin{itemize}
    \item \textbf{Pre-processing of the MS COCO dataset:} The extraction of Sentence-BERT embeddings and the subsequent application of the k-means clustering algorithm across the entire corpus of MS COCO captions require approximately 48 hours.
    
    \item \textbf{Generative processes:} The generative processes employing both the SDXL and GLIDE methods demand a total time of approximately 120 hours.
    
    \item \textbf{Preliminary relevance judgments:} The creation of initial relevance judgments for the retrieval task takes 72 hours.
    
    \item \textbf{Model fine-tuning:} The cumulative time spent on fine-tuning all predictors involved in our study amounts to 50 compute hours.
\end{itemize}

These estimates are based on the computational resources and configurations described in Section \ref{sec_resources}.

\section{Dataset Documentation}
\label{sec_docu}

\subsection{Documentation Framework}
The dataset is documented using the Data Card framework, which provides a comprehensive overview of its content, collection methods, and intended uses. The structure is as follows:

\begin{itemize}
\item \textbf{Dataset Overview:} General information about the dataset, including size, number of instances, and collected human labels.
\item \textbf{Content Description:} Detailed description of the data points, including relevant features and formats.
\item \textbf{Typical Data Point:} Example of a typical data entry.
\item \textbf{Dataset Structure:} Explanation of the dataset's organization, including file and folder descriptions.
\item \textbf{Provenance:} Information on data collection methods and maintenance status.
\item \textbf{Licensing:} Details about the dataset's license and usage terms.
\end{itemize}

\subsection{Dataset Overview}
The dataset does not contain sensitive data about people and includes original images from the MS COCO dataset. The dataset snapshot is as follows:

\begin{itemize}
\item \textbf{Size:} 34 GB
\item \textbf{Query/Prompt Instances:} 10,200
\item \textbf{Generated Image Instances:} 40,800
\item \textbf{Human Labels:} 1,589,055
\end{itemize}

\subsection{Dataset Format and Preservation}
\label{sec_format}

The dataset utilizes widely recognized open data formats. Annotations are provided in CSV format, while images are in standard image formats (PNG). Detailed instructions on reading and using the dataset are provided in the repository.

\subsection{Structured Metadata}
\label{sec
}
To enhance the discoverability and organization of our dataset, structured metadata is included using Web standards (schema.org). This metadata is encapsulated in a dataset.json file within our repository.

\subsection{Content Description}
Each data point includes the following features:

\begin{itemize}
\item \textbf{id:} Number, ID of the query in MS COCO / DrawBench.
\item \textbf{image\_id:} Number, ID of the image in MS COCO.

\item \textbf{best\_caption:} String, text containing selected prompt.
\item \textbf{blip2\_rr:} Float, reciprocal rank for query using BLIP-2 retrieval method.
\item \textbf{clip\_rr:} Float, reciprocal rank for query using CLIP retrieval method.
\item \textbf{blip2\_pk:} Float, precision@$10$ for the query using BLIP-2 retrieval method.
\item \textbf{clip\_pk:} Float, precision@$10$ for the query using CLIP retrieval method.
\item \textbf{glide\_score:} Human annotated generative score for the GLIDE model.
\item \textbf{sdxl\_score:} Human annotated generative score for the SDXL model.
\end{itemize}

\subsection{Typical Data Point}

\begin{table}[t]
\centering
\setlength\tabcolsep{3.5pt}
\small{
\begin{tabular}{|l|c|}
\hline
\textbf{Column Name} & \textbf{Value} \\ 
\hline
\hline
id & 319365 \\ 
image\_id & 363951 \\ 
\hline
best\_caption & Black and white of windsurfers on a lake. \\ 
\hline
blip2\_rr & 1.0 \\ 
clip\_rr & 1.0 \\ 
\hline
blip2\_pk & 0.1 \\ 
clip\_pk & 0.1 \\ 
\hline
glide\_score & 0.5 \\ 
sdxl\_score & 2.0 \\ 
\hline
\end{tabular}
}
\caption{Example of a typical data point.}
\label{tab:typical_data_point}
\end{table}

A typical data point is shown in Table \ref{tab:typical_data_point}.

\subsection{Dataset Structure}
The dataset folder structure can be viewed in the official repository:
\begin{itemize}
\item \textbf{Dataset Files:} CSV files for training, validation, and test splits containing MS COCO image IDs, P@$10$/RR scores for retrieval, and HBPP scores for the generative setting.
\item \textbf{Image Folder:} Contains the SDXL/GLIDE generated images alongside the original MS COCO images.
\end{itemize}
The folder structure is:
\begin{itemize}
\item \textbf{Dataset Files:} 
\begin{lcverbatim}
\dataset
    \ train.csv
    \ validation.csv
    \ test.csv
\end{lcverbatim}
\item \textbf{Image Folder:} 
\begin{lcverbatim}
\images
    \{IMG_ID}
        \image_4.png
        \image_5.png
        \image_6.png
        \image_7.png
        \image_8.png
\end{lcverbatim}
\end{itemize}

The structure of the additional resources is explained in extenso in the official repository.



\section{Maintenance and Support}

\subsection{Maintenance}
Although there is no plan to make new versions available in the future, this dataset will be actively maintained by the authors, including but not limited to updates to the data.

\subsection{Support}
We commit to maintaining the dataset and providing support through the following channels:

\begin{itemize}
\item Official Github repository ticketing system.
\item Direct contact via email at:\\ \texttt{eduardgabriel.poe@gmail.com}.
\end{itemize}

\section{Licensing and Responsibility Statement}
\label{sec_license}

We release our dataset, which includes annotations alongside images created with generative models, under the CC BY 4.0 license. We also acknowledge the license offered by the original authors of the MS COCO dataset annotations (CC BY 4.0) and the Flickr Terms of Use for the images, as detailed at \url{https://cocodataset.org/#termsofuse} and \url{https://www.flickr.com/creativecommons/}.

In the event that it is determined that we have violated any rights or licenses associated with the used resources, we take full responsibility and guarantee our cooperation in resolving any such issues with any affected third parties. Potential resolutions will include, as appropriate, the modification, substitution, or deletion of data or code that infringe on copyrights or licenses.

\section{Intended Uses}
\label{sec_uses}

This dataset is intended for use in either commercial or research and development within the domains of machine learning, computer vision, query performance prediction, and prompt performance prediction. It is designed to facilitate the training, validation, and testing of models for these applications.

\end{document}